\documentclass[sigconf]{acmart}
\usepackage{bm}
\usepackage{subcaption}

\def\BibTeX{{\rm B\kern-.05em{\sc i\kern-.025em b}\kern-.08emT\kern-.1667em\lower.7ex\hbox{E}\kern-.125emX}}

\begin{document}

%
\title[Towards Self-similarity Consistency and Feature Discrimination for Domain Adaptation]{Towards Self-similarity Consistency and Feature Discrimination for Unsupervised Domain Adaptation}



\author{Chao Chen}
\authornote{This work was done when the author was visiting Alibaba DAMO Academy as a research intern}
\affiliation{%
  \institution{Zhejiang University}
}
\email{chench@zju.edu.cn}

\author{Zhihang Fu}
\affiliation{%
  \institution{Alibaba Group}
}
\email{zhihang.fzh@alibaba-inc.com}

\author{Zhihong Chen}
\affiliation{%
 \institution{Zhejiang University}
}
\email{zhihongchen@zju.edu.cn}

\author{Zhaowei Cheng}
\affiliation{%
 \institution{Zhejiang University}
}
\email{chengzhaowei@zju.edu.cn}

\author{Xinyu Jin}
\authornote{Corresponding authors}
\affiliation{%
  \institution{Zhejiang University}
}
\email{jinxy@zju.edu.cn}

\author{Xian-Sheng Hua}
\authornotemark[2]
\affiliation{\institution{Alibaba Group}}
\email{huaxiansheng@gmail.com}
%

%
\begin{abstract}
Recent advances in unsupervised domain adaptation mainly focus on learning shared representations by global distribution alignment without considering class information across domains. The neglect of class information, however, may lead to partial alignment (or even misalignment) and poor generalization performance. For comprehensive alignment, we argue that the similarities across different features in the source domain should be consistent with that of in the target domain. Based on this assumption, we propose a new domain discrepancy metric, i.e., Self-similarity Consistency (\textbf{SSC}), to enforce the feature structure being consistent across domains. The renowned correlation alignment (CORAL) is proven to be a special case, and a sub-optimal measure of our proposed SSC. Furthermore, we also propose to mitigate the side effect of the partial alignment and misalignment by incorporating the discriminative information of the deep representations. Specifically, an embarrassingly simple and effective feature norm constraint is exploited to enlarge the discrepancy of inter-class samples. It relieves the requirements of strict alignment when performing adaptation, therefore improving the adaptation performance significantly. Extensive experiments on visual domain adaptation tasks demonstrate the effectiveness of our proposed SSC metric and feature discrimination approach.
\end{abstract}

%


%

%


%
\maketitle

\section{Introduction}
Convolutional neural networks (CNNs) have shown promising results on supervised learning tasks. However, the generalization ability of the learned model may degrade severely when applied to other related but different domains. It is often expensive or impractical to annotate massive samples on the coming new domains. Domain adaptation which aims to utilize labeled samples from source domain to annotate the target domain samples has, therefore, emerged as a new learning framework to address this problem \cite{csurka2017domain,wang2018deep}. In this paper, we mainly focus on the unsupervised domain adaptation (UDA) problems. Recent advances in UDA show satisfactory performance with deep CNNs. Among them, the most successful methods encourage similarities between the latent deep representations across different domains. The similarities are often maximized by some measure of discrepancy metrics \cite{tzeng2014deep,long2015learning,sun2016deep,long2017deep,zellinger2017central} or adversarial training \cite{ganin2016domain,tzeng2017adversarial,hoffman2018cycada}.

\begin{figure}[!t]
\begin{center}
\includegraphics[width=1.0\linewidth]{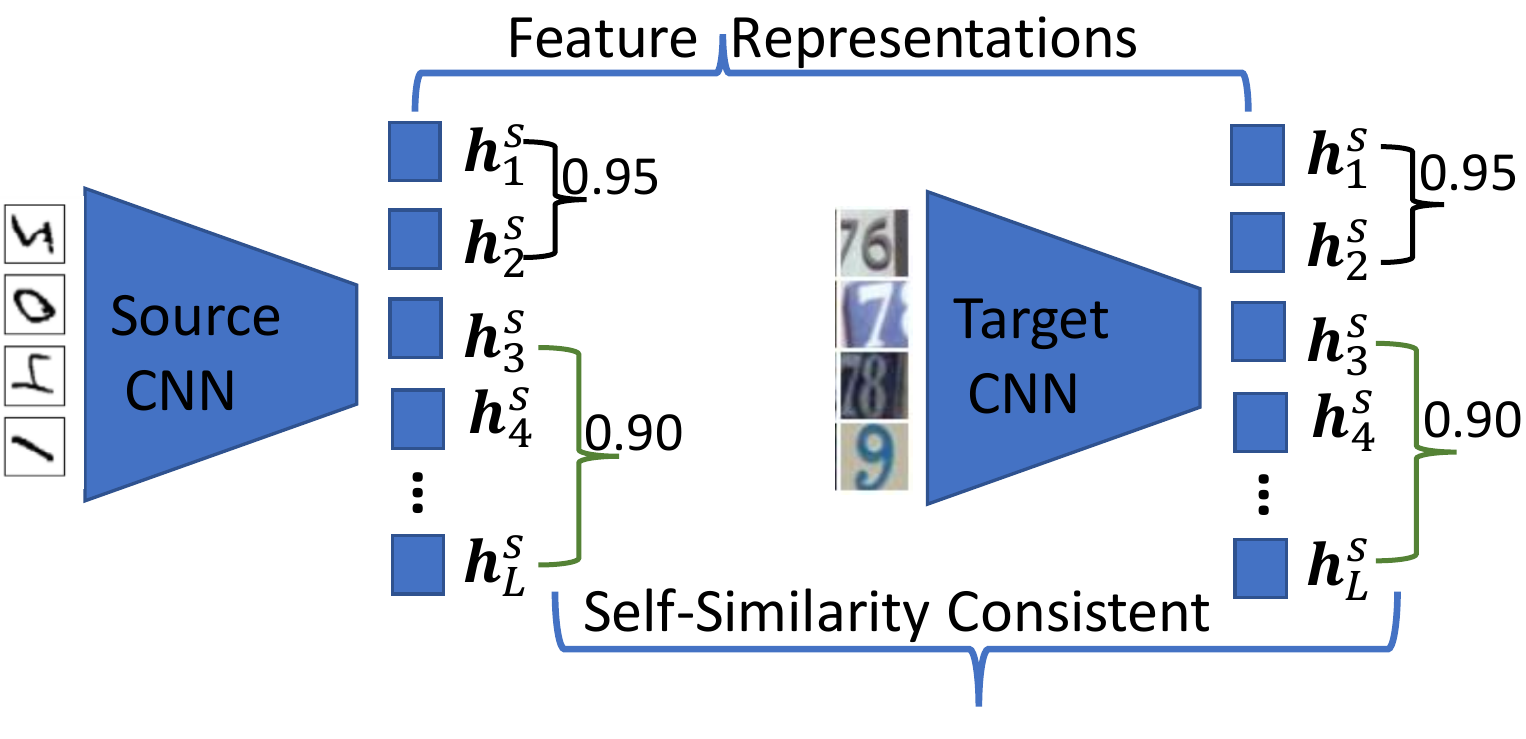}
\end{center}
   \caption{We propose a self-similarity consistency measure, which enforces the similarity/distance across different features to be identical in the source and target domains.}
\label{fig1}
\end{figure}

Maximum Mean Discrepancy(MMD) and Correlation Alignment (CORAL) are two most commonly used metrics to measure the distribution discrepancy. A great deal of methods based on the MMD \cite{tzeng2014deep,long2015learning,long2017deep,kang2019contrastive} and CORAL \cite{sun2016return,sun2016deep,zhang2018unsupervised} have achieved considerable adaptation performance.  However, the MMD-based methods suffer from a very critical limitation. They only align the global distribution statistics without any semantic information, which may even lead to negative transfer \cite{hoffman2018cycada,xie2018learning}. For example, the features of digit "2" in the source domain may be aligned well with the features of digit "3" in the target domain \cite{hoffman2018cycada}. Although, there are some researches attempting to alleviate this problem by leveraging the pseudo-label \cite{saito2017asymmetric,xie2018learning}, this problem has not been solved well due to lack of target label information. The prior adversarial adaptation methods also suffer from this limitation, as the discriminator only guarantees the global alignment of domain statistics lacking crucial semantic information for each category. Several recent works also attempt to address this problem by pixel-level domain alignment \cite{bousmalis2017unsupervised,hoffman2018cycada}, which performs image-to-image transformation for domain adaptation.
\begin{figure}[t!]
\centering
\includegraphics[width=1.0\linewidth]{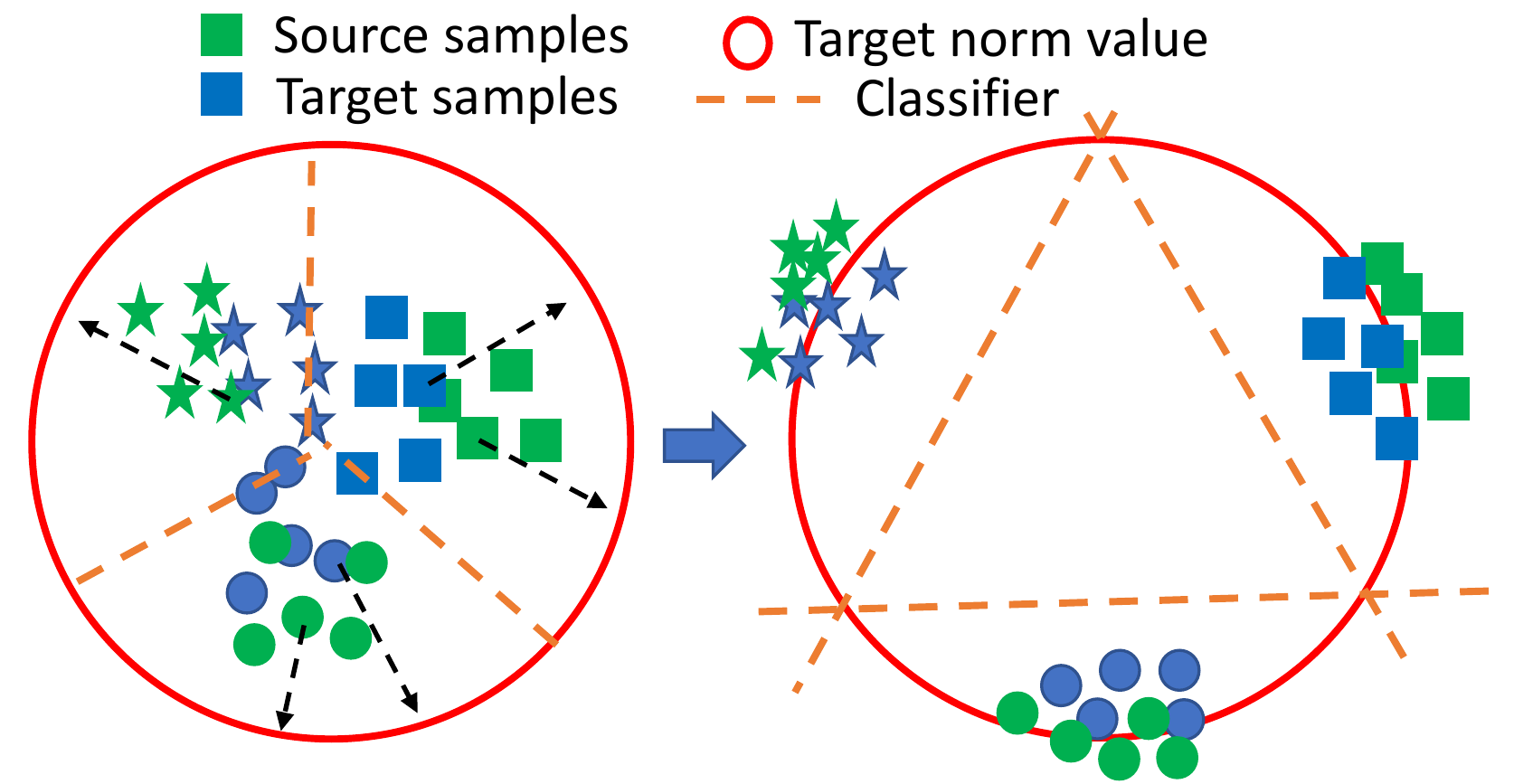}
\caption{Enforce feature discrimination by the feature norm constraint. \textbf{Left:} With the previous domain alignment methods, the target domain samples distributed in the edge of the cluster are prone to be misclassified. \textbf{Right:} With the feature discrimination loss, the norm of the deep features will be enlarged to be close to the target norm value, which make the inter-class samples more separable. (Best viewed in color)}
\label{fig2}
\end{figure}

\textbf{Motivation of SSC} In this paper, we propose a new self-similarity consistency (\textbf{SSC}) metric to measure the domain discrepancy. As illustrated in Fig. \ref{fig1}, an intuitive motivation is that if the source domain and the target domain are well aligned, the similarity (or distance) across different features in the source domain should be consistent with the similarity in the target domain. For instance, if the distance between $i$-th feature and $j$-th feature in the source domain is $d(\bm{h}_s^i,\bm{h}_s^j)=0.95$, then in the target domain this relationship should be also satisfied, i.e., $d(\bm{h}_t^i,\bm{h}_t^j)=0.95$. From this perspective, we propose  a SSC constraint to match the source and target domains and we demonstrate that the Correlation Alignment (CORAL) is only a special case, and a sub-optimal measure of the proposed SSC. Compared with the MMD-based methods which align the centroid of all the samples between the source and target domains, our proposed SSC performing structure similarity constraint of the deep features would be less susceptible to misalignment.

\textbf{Motivation of Feature Discrimination} Existing alignment methods can only reduce, but not remove domain discrepancy, which we call partial alignment. Therefore, the samples distributed in the edge of the cluster are prone to be misclassified. From this perspective, one intuitive approach to improve the adaptation performance is to enforce the deep features across domains with better intra-class compactness and inter-class separability \cite{chen2018joint}. As illustrated in Fig. \ref{fig2}, we propose an embarrassingly simple, but extremely effective method to enhance the discrimination of the deep features. Specifically, a feature norm constraint is employed to enforce the deep features to be scaled to a given larger hypersphere (feature norm). In this way, since the large margins between different classes, it is expected to get satisfactory adaptation performance even the source and target domains are not perfectly aligned (partially aligned). Besides, the possibility of misalignment can also be reduced due to the high discrimination of the deep features.

For simplicity, we denote our proposal "Towards \textbf{S}elf-similarity Consistency and Feature \textbf{D}iscrimination for Unsupervised \textbf{D}omain \textbf{A}daptation" as \textbf{SDDA}. The main contributions of this work can be concluded as: (1) We introduce a new metric self-similarity consistency (SSC) to measure the domain discrepancy which performs better than previous metrics on most of transfer tasks, and we demonstrate that CORAL can be viewed as a special case of SSC. (2) From the perspective of feature discrimination, we propose an embarrassingly simple approach to enlarge the separability of inter-class samples, which can improve the adaptation performance significantly.

\section{Related Work}
\textbf{Domain-Invariant Feature Learning.} Deep domain adaptation has provided promising performance for visual applications. Among them, the typical deep domain adaptation methods follow the Siamese CNN architectures with two streams \cite{sun2016deep,long2017deep,rozantsev2018beyond,chen2018joint}, representing the source model and the target model respectively. A practical way to perform domain adaptation is to minimize the domain discrepancy to obtain domain-invariant features. There has been a large body of work pays attention to learn the domain-invariant representations by some measures of domain discrepancy. The most widely used discrepancy metrics including MMD \cite{tzeng2014deep,long2015learning,long2017deep,kang2019contrastive}, CORAL \cite{sun2016return,sun2016deep,zhang2018unsupervised,morerio2018} and Central Moment Discrepancy (CMD) \cite{zellinger2017central}. Specifically, Long et al. proposed DAN \cite {long2015learning} and JAN \cite{long2017deep} which perform domain matching via multi-kernel MMD or a joint MMD criteria in multiple domain-specific layers across domains. Sun et al. proposed the correlation alignment (CORAL) \cite{sun2016return,sun2016deep} to align the second order statistics of the source and target distributions. Some recent work also extended the CORAL to mapped CORAL (MCA) \cite{zhang2018unsupervised} and logCORAL \cite{morerio2017correlation,morerio2018}. Besides, CMD \cite{zellinger2017central} which aligns the central moment of each order across domains is also an effective approach for domain alignment. Another fruitful line of work to learn the domain-invariant features is through adversarial training \cite{ganin2016domain,tzeng2017adversarial}, which encourages domain confusion by a domain adversarial objective whereby a discriminator (domain classifier) is trained to distinguish between the source and target representations. Also, recent work performing pixel-level adaptation by image-to-image transformation \cite{zhu2017unpaired,murez2018image,hoffman2018cycada} has achieved satisfactory performance and obtained much attention, which is also widely used for cross-domain segmentation \cite{huo2018synseg,sankaranarayanan2018learning} and person re-identification \cite{deng2018image}. In this work, we propose a novel self-similarity consistency (SSC) metric, which exploits the structure similarity of the feature space for domain matching, instead of aligning the global statistics of all the samples across domains.

\subsection{Discriminative Feature Learning.}
 Recently, there is a trend to improve the performance of CNN with discriminative feature learning, especially in the field of face recognition \cite{wen2016discriminative,zheng2018ring,liu2016large}, and person re-identification \cite{wang2018learning}. Wen et al. proposed the Center Loss \cite{wen2016discriminative} to learn the discriminative features by penalizing the distance of each sample to its corresponding class center. Liu et al. proposed the  large margin softmax (L-Softmax) \cite{liu2016large} by enforcing angular constrains to improve the feature discrimination. Besides, some recent work also improved the domain adaptation methods by incorporating the discriminative feature learning \cite{lu2017two,lu2018embarrassingly,li2018domain,kang2019contrastive,chen2018joint}. Chen et al. \cite{chen2018joint} proposed joint domain alignment and discriminative feature learning (JDDA), where an instance-based discriminative feature learning method and a center-based discriminative feature learning method are proposed to guarantee the domain invariant features with better intra-class compactness and inter-class separability. Kang et al. \cite{kang2019contrastive} proposed the Contrastive Adaptation Network (CAN) which explicitly models the inter-class domain discrepancy and inter-class domain discrepancy by revisiting the MMD. In this paper, we propose an elegant feature norm constraint for discriminative feature learning.

\begin{figure}[t!]
\centering
\includegraphics[width=1.0\linewidth]{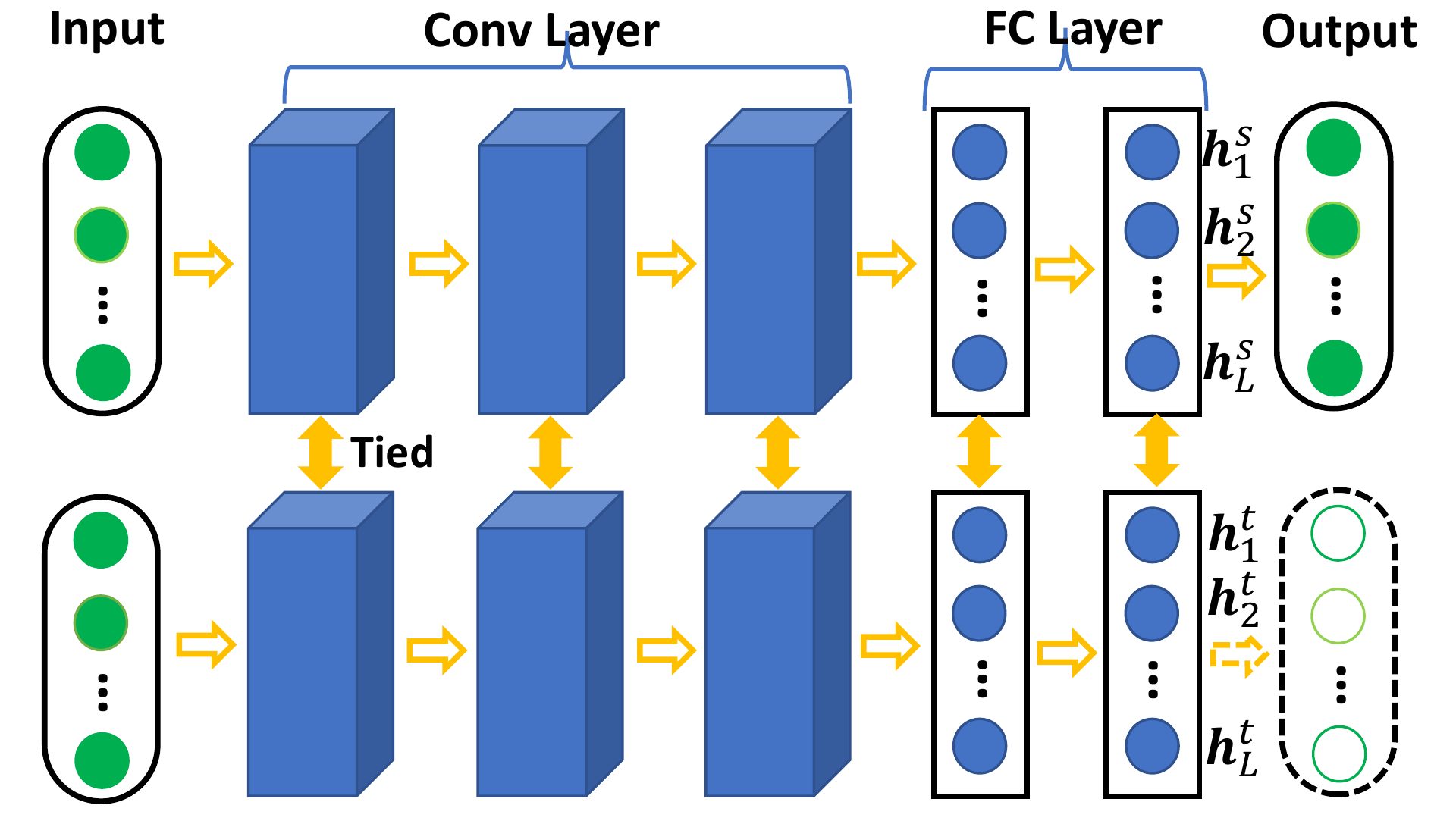}
\caption{Two-stream CNNs with shared parameters are adopted for unsupervised deep domain adaptation. The first stream operates the source data and the second stream operates the target data. The last FC layer (the input of the output layer) is used as the adapted layer. (Best viewed in color)}
\label{fig3}
\end{figure}

\section{Methodology}
In this work, we consider the unsupervised domain adaptation problem. Let $\mathcal{D}_s=\{\bm{x}_s^i,y_s^i\}_{i=1}^{n_s}$ denotes the source domain with $n_s$ labeled samples and $\mathcal{D}_t=\{\bm{x}_t^i\}_{i=1}^{n_t}$ denotes the target domain with $n_t$ unlabeled samples. We aim to train a cross-domain CNN classifier $\phi_{\bm\theta}(x)$ which can minimize the target risks $\epsilon_t=\mathbb{E}_{\bm{x}\in\mathcal{D}_t}[\phi_{\bm\theta}(x)\neq \bm{y}_t]$ with labeled source domain samples and unlabeled target domain samples. Here $\phi_{\bm\theta}(x)$ denotes the outputs of the deep neural networks, $\bm\theta$ denotes the model parameter to be learned. Following \cite{chen2018joint,long2017deep,sun2016deep}, we adopt the two-stream CNNs architecture for unsupervised deep domain adaptation.  As shown in Fig. \ref{fig3}, the two streams share the same parameters (tied weights), operating the source and target domain samples respectively. And we perform the domain alignment in the last full-connected (FC) layer \cite{sun2016deep,chen2018joint}. In this paper, we minimize the domain discrepancy by the proposed SSC metric. Besides, we also optimize the domain-invariant feature representations with better intra-class compactness and inter-class separability. The overall objective can be given as
\begin{equation}\label{eq1}
\begin{split}
\mathcal{L}(\bm\theta|\mathbf{X}_s,\mathbf{Y}_s,\mathbf{X}_t)=&\mathcal{L}_s+\lambda_{ssc}\mathcal{L}_{ssc}+\lambda_{intra}\mathcal{L}_{d}^{intra}\\
&+\lambda_{inter}\mathcal{L}_{d}^{inter}
\end{split}
\end{equation}
\begin{equation}\label{eq2}
\mathcal{L}_s=\dfrac{1}{n_s}\sum\limits_{i=1}^{n_s}J(\phi_{\bm\theta}(x_i^s), \bm{y}_i^s)
\end{equation}
where $\mathcal{L}_s$ represents the standard classification loss in the source domain, $J(\cdot,\cdot)$ represents the cross-entropy loss function. $\mathcal{L}_{ssc}$ represents the domain discrepancy loss measured by the proposed SSC metric. $\mathcal{L}_{d}^{intra}$ and $\mathcal{L}_{d}^{inter}$ denote intra-class compactness loss and inter-class separability loss, respectively. $\lambda_{ssc}$, $\lambda_{intra}$ and $\lambda_{inter}$ are three trade-off parameters which balance the contributions of the domain discrepancy loss and the feature discrimination loss.

\begin{figure}[t!]
\centering
\includegraphics[width=1.0\linewidth]{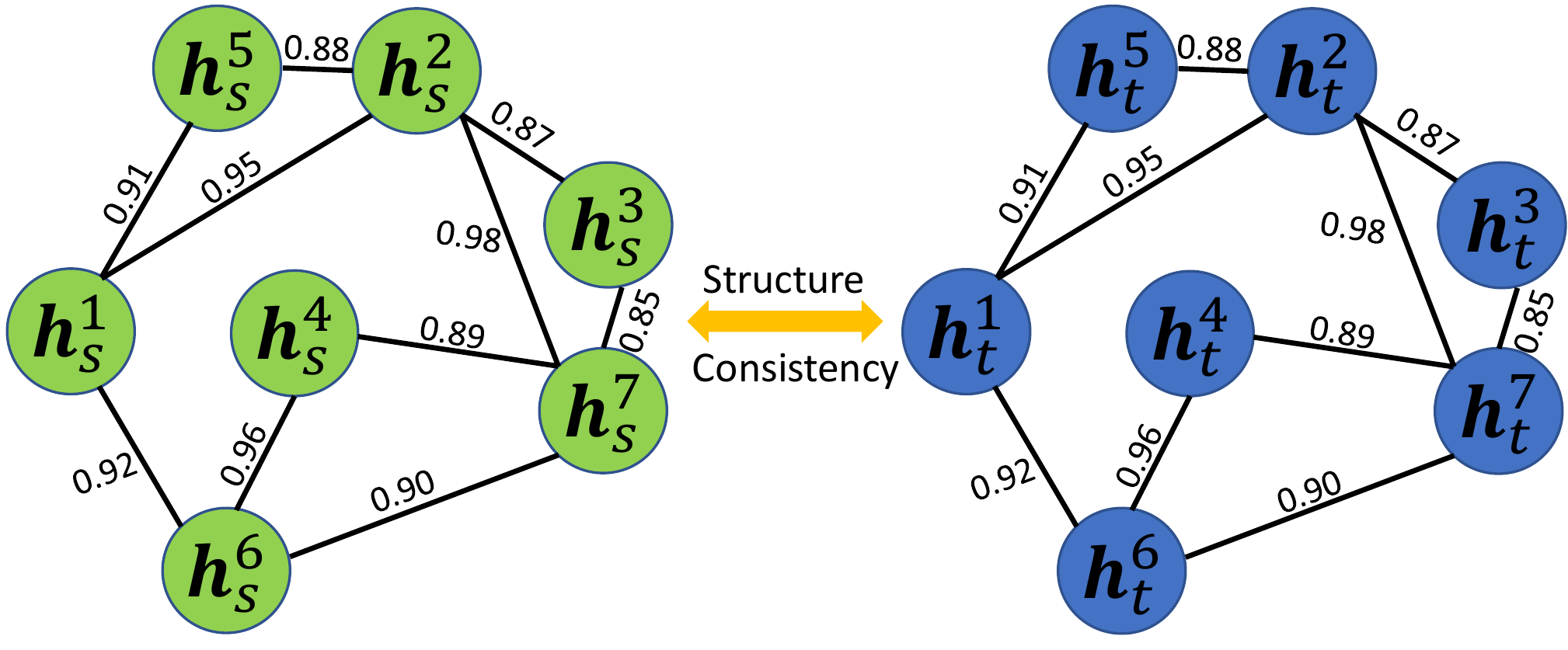}
\caption{Perform self-similarity consistency constraint to minimize the domain discrepancy. The SSC metric matches the pairwise similarity across the deep features in the source and target domain, which ensures the structure of the source feature space be consistent with the structure of the target feature space. (Best viewed in color)}
\label{fig4}
\end{figure}

\subsection{Self-similarity Consistency}
To minimize the domain discrepancy, we propose a novel self-similarity consistency metric to match the structure of the feature space across domains. Let $\mathbf{H}_s=[\widetilde{\bm{h}}^s_1,\widetilde{\bm{h}}^s_2,\cdots,\widetilde{\bm{h}}^s_L]_{b\times L}$ and $\mathbf{H}_t=[\widetilde{\bm{h}}^t_1,\widetilde{\bm{h}}^t_2,\cdots,\widetilde{\bm{h}}^t_L]_{b\times L}$ denote the centralized source outputs and target outputs in the last full-connected (FC) layer (as shown in Fig. \ref{fig3}). Here, $\widetilde{\bm{h}}^{s(t)}_i=\bm{h}^{s(t)}_i-\overline{\bm{h}}^{s(t)}$, $L$ denotes the number of hidden nodes in the last FC layer, $b$ denotes the batch size during the training stage. As can be seen in Fig. \ref{fig4}, the motivation of the SSC metric is that the similarity or distance across different features in the source domain should be consistent with the corresponding similarity or distance in the target domain, i.e., $\bm{D}^s_{ij}=\bm{D}^t_{ij}$, for example, $\bm{D}^s_{12}=\bm{D}^t_{12}=0.95$. Then, the generic SSC loss can be defined as
\begin{equation}\label{eq5}
\mathcal{L}_{ssc}=\Vert\bm{D}^s-\bm{D}^t\Vert^2_F
\end{equation}
Here $\bm{D}_s\in\mathbb{R}^{L\times L}$ and $\bm{D}_t\in\mathbb{R}^{L\times L}$ represent the self-similarity matrix across different features in the source and target domains, and $\bm{D}^{s(t)}_{ij}$ denotes the similarity between the $i$-th feature and $j$-th feature in the source (target) domain. The commonly used similarity can be defined as:
\begin{itemize}
\item Dot-product Similarity $sim(\bm{x},\bm{y})=\bm{x}\cdot \bm{y}$
\item Euclidean Distance $sim(\bm{x},\bm{y})=\Vert\bm{x}-\bm{y}\Vert_2$
\item Cosine Similarity $sim(\bm{x},\bm{y})=\tfrac{\bm{x}\cdot\bm{y}}{\Vert\bm{x}\Vert\cdot\Vert\bm{y}\Vert}$
\end{itemize}
\textbf{Rethinking the Correlation Alignment (CORAL)} The renowned correlation alignment (CORAL) diminishes the domain discrepancy by aligning the covariance of the source and target domains. It is one of the most widely used metric for domain adaptation, which can be expressed as
\begin{equation}\label{Coral}
\mathcal{L}_{coral}=\tfrac{1}{L^2}\Vert \bm{C}^s-\bm{C}^t\Vert_F^2
\end{equation}
Here, $\bm{C}^s=\mathbf{H}_s^{\top}\mathbf{H}_s$ and $\bm{C}^t=\mathbf{H}_t^{\top}\mathbf{H}_t$ denote the source feature covariance and target feature covariance, respectively. And $\bm{C}^{s(t)}_{ij}=\widetilde{\bm{h}}^{s(t)}_i\cdot\widetilde{\bm{h}}^{s(t)}_j$ can be regarded as the dot-product similarity between $i$-th feature and $j$-th feature. In this respect, the CORAL can be viewed as a special case of our proposed SSC metric, in which the dot-product similarity is adopted.

\noindent\textbf{Heat Kernel Similarity} In this work, we adopt the heat kernel function to measure the similarity across different features. i.e.,
\begin{equation}\label{eq3}
\bm{D}^s_{ij}=exp(-\gamma\Vert\widetilde{\bm{h}}^s_i-\widetilde{\bm{h}}^s_j\Vert_2)
\end{equation}
\begin{equation}\label{eq4}
\bm{D}^t_{ij}=exp(-\gamma\Vert\widetilde{\bm{h}}^t_i-\widetilde{\bm{h}}^t_j\Vert_2)
\end{equation}
where $\gamma$ is the bandwidth to adjust the influence of single pairwise similarity. Note that the heat kernel similarity has the same expression as the Gaussian kernel (or RBF kernel) $k(\bm{x},\bm{z})=exp(-\tfrac{\Vert\bm{x}-\bm{z}\Vert^2}{2\delta^2})$.

\noindent\textbf{Kernel Embeding Perspective} Suppose we have a feature space $\mathbf{H}$ with $L$ features, and a feature map $\Phi: \mathbf{H}\rightarrow\mathbf{F}$, where $\mathbf{F}$ is an embedding space. The kernel matrix is defined as $\mathbf{K}=[k(\bm{h}_i,\bm{h}_j)]_{i,j=1}^L$, where $\mathbf{K}_{ij}=k(\bm{h}_i,\bm{h}_j)=\Phi(\bm{h}_i)\cdot\Phi_(\bm{h}_j)$. For the linear kernel embedding $\mathbf{K}_{ij}=k(\bm{h}_i,\bm{h}_j)=\bm{h}_i\cdot\bm{h}_j$, and for the RBF kernel embedding (or Guassian kernel embedding) $\mathbf{K}_{ij}=k(\bm{h}_i,\bm{h}_j)=exp(-\gamma\Vert\bm{h}_i-\bm{h}_j\Vert_2)$.  From this perspective, CORAL can be regarded as a kernel embedding method which aligns the source and target features in the linear kernel embedding, while our proposed SSC aligns the features in the RBF kernel embedding. Both theoretical analysis and experimental results show that our method is superior to CORAL, and CORAL can be viewed as a form of the self-similarity consistency metric.

\noindent\textbf{Relationship to Mean Map Kernel} {\it Mean map kernel} (MMK) \cite{muandet2012learning,shan2016randomized} measures the similarity between two distributions, which can be formulated as
\begin{equation}
K^{MMK}(\mathcal{X}_s,\mathcal{X}_t)=\frac{1}{|\mathcal{X}_s||\mathcal{X}_t|}\sum_{\bm{x}\in \mathcal{X}_s}\sum_{\bm{z}\in \mathcal{X}_t}k(\bm{x},\bm{z})
\end{equation}
where $k(\bm\cdot,\bm\cdot)$ is a kernel function and $|\mathcal{X}_{s(t)}|$ denotes the number of samples in $\mathcal{X}_{s(t)}$. In terms of domain adaption, MMK can be further transformed into a mutual-similarity maximization (MSM) metric when we replace the kernel function and the distributions $\mathcal{X}_s$, $\mathcal{X}_t$ with similarity function and centralized features, i.e.,
\begin{equation}
\mathcal{L}_{MSM}=-\frac{1}{L^2}\sum_{i=1}^{L}\sum_{j=1}^{L}sim(\widetilde{\bm{h}}_i^s,\widetilde{\bm{h}}_j^t)
\end{equation}
Similar to SSC, MSM is also an effective metric for domain matching. What sets SSC apart from MSM is that SSC aligns the self-similarity relationship across domains for domain matching, while MSM aggregates the pairwise similarity over two feature sets to measure the similarity of two distributions. In this work, we focus on the effectiveness of our proposed SSC using the heat kernel similarity. Investigation about mutual-similarity maximization is beyond the scope of this paper.

\subsection{Feature Discrimination}
Recently, there is a line of work improving the adaptation performance by pursuing the better intra-class compactness and inter-class separability of the domain-invariant features \cite{lu2017two,lu2018embarrassingly,li2018domain,kang2019contrastive,chen2018joint}. However, these methods are quite garish, and most of them are based on traditional classifier \cite{lu2017two,lu2018embarrassingly,li2018domain}. JDDA \cite{chen2018joint} provides an elegant approach to learn discriminative features for deep domain adaptation, but the inter-class separability is not well guaranteed. In this paper, we propose an extremely intuitive and effective approach, namely feature norm constraint, to ensure the shared representations with better inter-class separability.

\textbf{Intra-class Compactness} To make the shared representations with better intra-class compactness, we follow \cite{chen2018joint} to penalize the distances between the deep features and their corresponding class center. Let $\phi_{\bm\theta_f}(\bm{x}^s_i)$ denotes the deep features of the source samples $\bm{x}^s_i$ in the last FC layer, $\bm{c}_{y_i}\in\mathbb{R}^L$ denotes the $y_i$-th class center of the deep features, $y_i=\{1,2,\cdots,k\}$ and $k$ is the number of class. Then, the intra-class compactness loss can be formulated as
\begin{equation}\label{eq6}
\mathcal{L}_d^{intra}=\sum_{i=1}^{n_s}\max(0,\Vert\phi_{\bm\theta_f}(\bm{x}^s_i)-\mathbf{c}_{y_i}\Vert_2^2-m)
\end{equation}
the intra-class compactness loss will enforce the distance between the deep features $\phi_{\bm\theta_f}(\bm{x}^s_i)$ and its corresponding center $\mathbf{c}_{y_i}$ no more than the given margin $m$. Note that the actual global centers should be calculated by averaging all the samples. However, since we perform update based on mini-batch, the centers might be misestimated because of the noisy samples. Therefore, we use the moving average of the samples as the global centers, which can be updated in each iteration as
\begin{equation}\label{Eq7}
\Delta\mathbf{c}_j=\dfrac{\sum_{i=1}^b\delta(y_i=j)(\phi_{\bm\theta_f}(\bm{x}^s_i))}{1+\sum_{i=1}^b\delta(y_i=j)}
\end{equation}
\begin{equation}\label{Eq8}
\mathbf{c}_j^{t+1}=\alpha\mathbf{c}_j^t+(1-\alpha)\Delta\mathbf{c}_j^t
\end{equation}
where $\alpha$ is the learning rate of the centers, $b$ denotes the batch size, $\delta(condition)=1$ if $condition$ is satisfied. For simplicity, we set the margin $m=0$ and $\alpha=0.5$ throughout the experiments. Note that the intra-class compactness constraint only penalizes the source domain compactness since the target samples lack of category information. Fortunately, we obverse that the visual representations learned by deep CNNs are fairly domain invariant, i.e, the source samples and target samples have similar distributions in the feature space. As a result, it is reasonable to make the source features more discriminative, such that the target features maximally aligned with the source domain will become discriminative automatically \cite{chen2018joint}.

\textbf{Inter-class Separabiliy} Recall that the feature norm represents the distance between the hypersphere origin and the feature vector. There should be a  better separability among the inter-class samples if the feature norm can be enlarged.  As can be seen in Fig. \ref{fig2}, we propose to maximize the inter-class discrepancy by penalizing the difference between the given target norm value and the feature norm in the shared feature space. In this way, the feature representations would be pulled away from the hypersphere origin, making the domain-invariant features more separable. The feature norm constraint loss can be defined as
\begin{equation}\label{eq9}
\begin{split}
\mathcal{L}_d^{inter}=&\sum_{\bm{x}_i\in\mathcal{D}_s}(R_{tn}-\Vert\phi_{\bm\theta_f}(\bm{x}^s_i)\Vert_2)^2 \\
&+\sum_{\bm{x}_i\in\mathcal{D}_t}(R_{tn}-\Vert\phi_{\bm\theta_f}(\bm{x}^t_i)\Vert_2)^2
\end{split}
\end{equation}
Here, $R_{tn}$ is the target norm constant that we would like the domain-invariant features to be scaled to. The gradient with respect to the input features can be calculated as
\begin{equation}\label{eq10}
\begin{aligned}
\frac{\partial{\mathcal{L}_d^{inter}}}{\partial\phi_{\bm\theta_f}(\bm{x}^s_i)}&=\frac{\partial(R_{tn}-\Vert\phi_{\bm\theta_f}(\bm{x}^s_i)\Vert_2)^2}{\partial\Vert\phi_{\bm\theta_f}(\bm{x}^s_i)\Vert_2}
\cdot\frac{\partial\Vert\phi_{\bm\theta_f}(\bm{x}^s_i)\Vert_2}{\partial\phi_{\bm\theta_f}(\bm{x}^s_i)} \\
&=-2(R_{tn}-\Vert\phi_{\bm\theta_f}(\bm{x}^s_i)\Vert_2)\cdot\frac{\phi_{\bm\theta_f}(\bm{x}^s_i)}{\Vert\phi_{\bm\theta_f}(\bm{x}^s_i)\Vert_2} \\
&=-2(\frac{R}{\Vert\phi_{\bm\theta_f}(\bm{x}^s_i)\Vert_2}-1)\phi_{\bm\theta_f}(\bm{x}^s_i)
\end{aligned}
\end{equation}
There are several advantages to perform the feature norm constraint. (1) The larger feature norm will guarantee the inter-class samples with better separability. (2) The norm constraint will enforce the features to distribute around the same hypersphere, which mitigates the domain discrepancy to a certain degree. (3) The large gaps across different classes can relieve the requirements of strict alignment across domains, and mitigate the side effect of partial alignment and misalignment.

\section{Experiments}
\subsection{Setup}
\noindent\textbf{Dataset}
We evaluate the performance of our proposed SDDA by comparing against several state-of-the-art deep domain adaptation methods on two public unsupervised visual adaptation datasets: digital recognition dataset and office-31 dataset. The digital recognition dataset includes five widely used benchmarks: MNIST, USPS, Street View House Numbers (SVHN), MNIST-M, and SYN (synthetic digits dataset). Following the experimental protocol of \cite{chen2018joint}, we evaluate our proposal across four adaptation shifts, including: \textbf{SVHN}$\rightarrow$\textbf{MNIST}, \textbf{MNIST}$\rightarrow$\textbf{MNIST-M}, \textbf{USPS}$\rightarrow$\textbf{MNIST} and \textbf{SYN}$\rightarrow$\textbf{MNIST}. Office-31 is another commonly used dataset for real-world domain adaptation scenario, which contains 31 categories acquired from the office environment in three distinct image domains: \textbf{A}mazon (product images download form amazon.com), \textbf{W}ebcam (low-resolution images taken by a webcam) and \textbf{D}slr (high-resolution images taken by a digital SLR camera). The office-31 dataset contains 4110 images in total, with 2817 images in \textbf{A} domain, 795 images in \textbf{W} domain and 498 images in \textbf{D} domain. We evaluate our method on all the six transfer tasks $\textbf{A}\rightarrow\textbf{W}$, $\textbf{D}\rightarrow\textbf{W}$, $\textbf{W}\rightarrow\textbf{D}$, $\textbf{A}\rightarrow\textbf{D}$, $\textbf{D}\rightarrow\textbf{A}$ and $\textbf{W}\rightarrow\textbf{A}$ as \cite{sun2016deep,long2017deep}.

\noindent\textbf{Baselines}
To evaluate the effectiveness of our proposed SDDA, we compare the SDDA with the following competing methods, which are most related to our work. \\
\noindent\textbf{Source Only}: As an empirical lower bound, we train a model with the source samples only, and test it directly with the target samples.\\
\noindent\textbf{DDC} \cite{tzeng2014deep}: DDC is the first method that maximizes domain invariance by MMD metric using two-streams CNNs. \\
\noindent\textbf{DAN} \cite{long2015learning}: DAN learns more transferable features by embedding deep features of all task-specific layers in a reproducing kernel Hilbert space (RKHSs) and matching domain distributions optimally using multi-kernel MMD. \\
\noindent\textbf{CORAL} \cite{sun2016deep}: Deep correlation alignment minimizes domain discrepancy by aligning the second-order statistics of the source and target distributions. \\
\noindent\textbf{DANN} \cite{ganin2016domain}: DANN is an adversarial representation learning approach that uses a domain classifier to learn features that are both discriminative and invariant to the change of domains. \\
\noindent\textbf{ADDA} \cite{tzeng2017adversarial}: ADDA uses a discriminative base model, unshared weights, and the standard GAN loss to learn a discriminative mapping of target images to the source feature space by fooling a domain discriminator. \\
\noindent\textbf{CMD} \cite{zellinger2017central}: CMD proposes to align the central moment of each order across domains for domain adaptation. \\
\noindent\textbf{CyCADA} \cite{hoffman2018cycada}: CyCADA is a pixel-level unsupervised domain adaptation method that unifies cycle-consistent image translation with adversarial adaptation methods. \\
\noindent\textbf{JDDA} \cite{chen2018joint}: JDDA is the first work to perform domain matching and discriminative feature learning jointly for deep domain adaptation.

\noindent\textbf{Implementation details.}
In our experiments on digit recognition dataset, we utilize the modified LeNet whereby a bottleneck layer with 64 hidden nodes is added before the output layer for domain matching. Since the image size is different across different domains, we resize all the images to $32\times32$ and convert the RGB images to grayscale. For the experiments on Office-31, we use ResNet-50 pretrained on ImageNet as our backbone networks. And the activation output of the last full connected layer is used as feature representation for domain matching. Due to the small samples size of Office-31 dataset, we only update the weights of the
full-connected layers (fc) as well as the final block (scale5/block3), and fix other parameters pretrained on ImageNet. Follow the standard protocol of \cite{long2015learning,sun2016deep,chen2018joint}, we use all the labeled source domain samples and all the unlabeled target domain samples for training. All the comparison methods are based on the same CNN architecture for a fair comparison, and all the model parameters are shared between the source and target domains.
\\\indent For MMD-based methods (DDC, DAN), we use a linear combination of 19 RBF kernels with the standard deviation parameters ranging from $10^{-6}$ to $10^{6}$. For DANN regularization, we add a GRL and then a domain classifier with one hidden layer of 100 nodes. When training with ADDA, our adversarial discriminator consists of 3 fully connected layers: two layers with 500 hidden units followed by the final discriminator output.

\begin{table*}[ht]
\centering
\caption{results (accuracy \%) on digital recognition dataset  for unsupervised domain adaptation based on modified LeNet}\label{tab:aStrangeTable}
\label{tab1}
\begin{tabular}{cccccc}
\toprule
Method& SVHN$\rightarrow$MNIST&MNIST$\rightarrow$MNIST-M&USPS$\rightarrow$MNIST&SYN$\rightarrow$MNIST&Avg\\
\midrule
Source Only &67.3$\pm$0.3&62.8$\pm$0.2&$66.4\pm$0.4&89.7$\pm$0.2&71.6\\
DDC \cite{tzeng2014deep}&71.9$\pm$0.4&78.4$\pm$0.1&75.8$\pm$0.3&89.9$\pm$0.2&79.0\\
DAN \cite{long2015learning}&79.5$\pm$0.3&$79.6\pm$0.2&$89.8\pm$0.2&75.2$\pm$0.1&81.0\\
DANN \cite{ganin2016domain}&70.6$\pm$0.2&76.7$\pm$0.4&76.6$\pm$0.3&90.2$\pm$0.2&78.5\\
CMD \cite{zellinger2017central}&86.5$\pm$0.3&85.5$\pm$0.2&86.3$\pm$0.4&96.1$\pm$0.2&88.6\\
ADDA \cite{tzeng2017adversarial}&72.3$\pm$0.2&$80.7\pm$0.3&92.1$\pm$0.2&96.3$\pm$0.4&85.4\\
CORAL \cite{sun2016deep}&89.5$\pm$0.2&81.6$\pm$0.2&96.5$\pm$0.3&96.5$\pm$0.2&91.0\\
CyCADA \cite{hoffman2018cycada}&92.8$\pm$0.1&\textbf{98.3}$\pm$0.2&97.4$\pm$0.3&97.5$\pm$0.1&\textbf{96.5}\\
JDDA \cite{chen2018joint} &94.2 $\pm$0.1&88.4$\pm$0.2&96.7$\pm$0.1&97.7$\pm$0.0&94.3\\
\midrule
\textbf{SDDA} (w/o FD) &94.2$\pm$0.2&82.2$\pm$0.2&96.5$\pm$0.2&97.3$\pm$0.1&92.6\\
\textbf{SDDA} ($\lambda_{inter}=0$) &94.9$\pm$0.1&88.9$\pm$0.2&96.9$\pm$0.1&97.6$\pm$0.0&94.6\\
\textbf{SDDA} ($\lambda_{intra}=0$) &96.5$\pm$0.1&87.9$\pm$0.3&\textbf{98.5}$\pm$0.1&\textbf{98.8}$\pm$0.0&95.4\\
\textbf{SDDA} (Full) &\textbf{97.3}$\pm$0.3&90.5$\pm$0.3&97.6$\pm$0.1&\textbf{98.8}$\pm$0.0&\textbf{96.1}\\
\bottomrule
\end{tabular}
\footnotesize \\ SDDA (w/o FD) indicates that the feature discrimination loss is not involved in the SDDA, i.e., $\lambda_{intra}=0$ and $\lambda_{inter}=0$.
\end{table*}

\begin{table*}[ht]
\centering
\caption{results (accuracy \%) on Office-31 dataset for unsupervised domain adaptation based on ResNet-50}\label{tab:aStrangeTable}
\label{tab2}
\begin{tabular}{cccccccc}
\toprule
Method& A$\rightarrow$W&D$\rightarrow$W&W$\rightarrow$D&A$\rightarrow$D&D$\rightarrow$A&W$\rightarrow$A&Avg\\
\midrule
Source Only &73.1$\pm$0.2&93.2$\pm$0.2&$98.8\pm$0.1&72.6$\pm$0.2&55.8$\pm$0.1&56.4$\pm$0.3&75.0\\
DDC \cite{tzeng2014deep}&74.4$\pm$0.3&94.0$\pm$0.1&98.2$\pm$0.1&74.6$\pm$0.4&56.4$\pm$0.1&56.9$\pm$0.1&75.8\\
DAN \cite{long2015learning}&78.3$\pm$0.3&$95.2\pm$0.2&$99.0\pm$0.1&75.2$\pm$0.2&58.9$\pm$0.2&64.2$\pm$0.3&78.5\\
DANN \cite{ganin2016domain}&73.6$\pm$0.3&94.5$\pm$0.1&99.5$\pm$0.1&74.4$\pm$0.5&57.2$\pm$0.1&60.8$\pm$0.2&76.7\\
CMD \cite{zellinger2017central}&76.9$\pm$0.4&94.6$\pm$0.3&99.2$\pm$0.2&75.4$\pm$0.4&56.8$\pm$0.1&61.9$\pm$0.2&77.5\\
CORAL \cite{sun2016deep}&79.3$\pm$0.3&94.3$\pm$0.2&99.4$\pm$0.2&74.8$\pm$0.1&56.4$\pm$0.2&63.4$\pm$0.2&78.0\\
CyCADA \cite{hoffman2018cycada}&82.2$\pm$0.3&94.6$\pm$0.2&99.7$\pm$0.1&78.7$\pm$0.1&60.5$\pm$0.2&67.8$\pm$0.2&80.6\\
JDDA\cite{chen2018joint}&82.6$\pm$0.4&95.2$\pm$0.2&99.7$\pm$0.0&79.8$\pm$0.1&57.4$\pm$0.0&66.7$\pm$0.2&80.2\\
\midrule
\textbf{SDDA} (w/o FD) &82.4$\pm$0.2&94.7$\pm$0.1&99.3$\pm$0.0&77.8$\pm$0.2&56.9$\pm$0.1&65.1$\pm$0.3&79.4\\
\textbf{SDDA} ($\lambda_{inter}=0$) &83.9$\pm$0.3&95.3$\pm$0.2&99.3$\pm$0.0&80.4$\pm$0.1&59.7$\pm$0.3&67.3$\pm$0.3&81.0\\
\textbf{SDDA} ($\lambda_{intra}=0$) &84.7$\pm$0.2&\textbf{99.1}$\pm$0.1&\textbf{99.8}$\pm$0.0&81.2$\pm$0.2&64.9$\pm$0.2&66.7$\pm$0.2&82.7\\
\textbf{SDDA} (Full) &\textbf{87.5}$\pm$0.3&\textbf{98.8}$\pm$0.0&\textbf{99.8}$\pm$0.0&\textbf{86.4}$\pm$0.3&\textbf{67.1}$\pm$0.2&\textbf{69.4}$\pm$0.3&\textbf{84.8}\\
\bottomrule
\end{tabular}
\end{table*}

All these methods are implemented via tensorflow and we use the Adam optimizer with the learning rate of 1e-4 to train the network. Regarding the optimal hyper-parameters, they are determined by applying multiple experiments using grid search strategy. The optimal hyper-parameters may distinct across different transfer tasks. Specifically, the trade-off parameters are selected as (or chosen from) $\lambda_{ssc}=1000$, $\lambda_{intra}\in\{0.001,0.003,0.01\}$ and $\lambda_{inter}\in\{0.00001,0.0001,0.0005\}$  throughout the experiments. For the digit recognition tasks, the hyper-parameter $\gamma$ in SSC is set as 0.001 and the target norm value is set as $R_{tn}=10$. For the experiments on office-31, the hyper-parameter $\gamma$ in SSC is set as 0.0001, and the target norm value is set as $R_{tn}=100$.  When implementing the comparison baselines, we follow the learning rate schedule described in \cite{long2015learning,long2017deep}, i.e., the adaptation factor $\lambda$ is gradually updated from 0 to 1 by a progressive schedule: $\lambda=\tfrac{2}{1+exp(-\mu p)-1}$ where $\mu$ is a constant set as 10 and $p$ is the training progress linearly changing from 0 to 1.

\subsection{Experimental results}
\textbf{Digit Recognition} Table \ref{tab1} shows the adaptation performance on four transfer tasks of digit recognition dataset based on the modified LeNet. As can be seen, all the domain adaptation methods outperform the source only (non-adapted) model by a large margin, while our proposed SDDA yields notable improvement over the comparison methods on most of the transfer tasks. In particular, our method improves the adaption performance significantly in the hard transfer tasks, such as, SVHN$\rightarrow$MNIST and MNIST$\rightarrow$MNIST-M. To our best knowledge, our approach achieves the highest classification accuracy in the transfer tasks SVHN$\rightarrow$MNIST over all the unsupervised domain adaptation methods. The SVHN is colored and some images contain multiple digits, while the MNIST is gray scale without messy background, thus this domain shift is a challenging adaptation scenario as well as a representative transfer task. Note that the MNIST-M was created by using each MNIST digit as a binary mask and inverting with it the colors of a background image randomly cropped from the Berkeley Segmentation Data Set \cite{arbelaez2011contour}. Therefore, the pixel-level domain adaptation methods, such as \cite{hoffman2018cycada,bousmalis2017unsupervised,murez2018image} can transfer the "MNIST-like" images to "MNIST-M-like" images easily. This is why CyCADA achieves much higher classification accuracy compared with other methods on the transfer task MNIST$\rightarrow$MNIST-M.

\begin{figure*}[!ht]
    \centering
  \begin{subfigure}[b]{0.248\textwidth}
    \includegraphics[width=\textwidth]{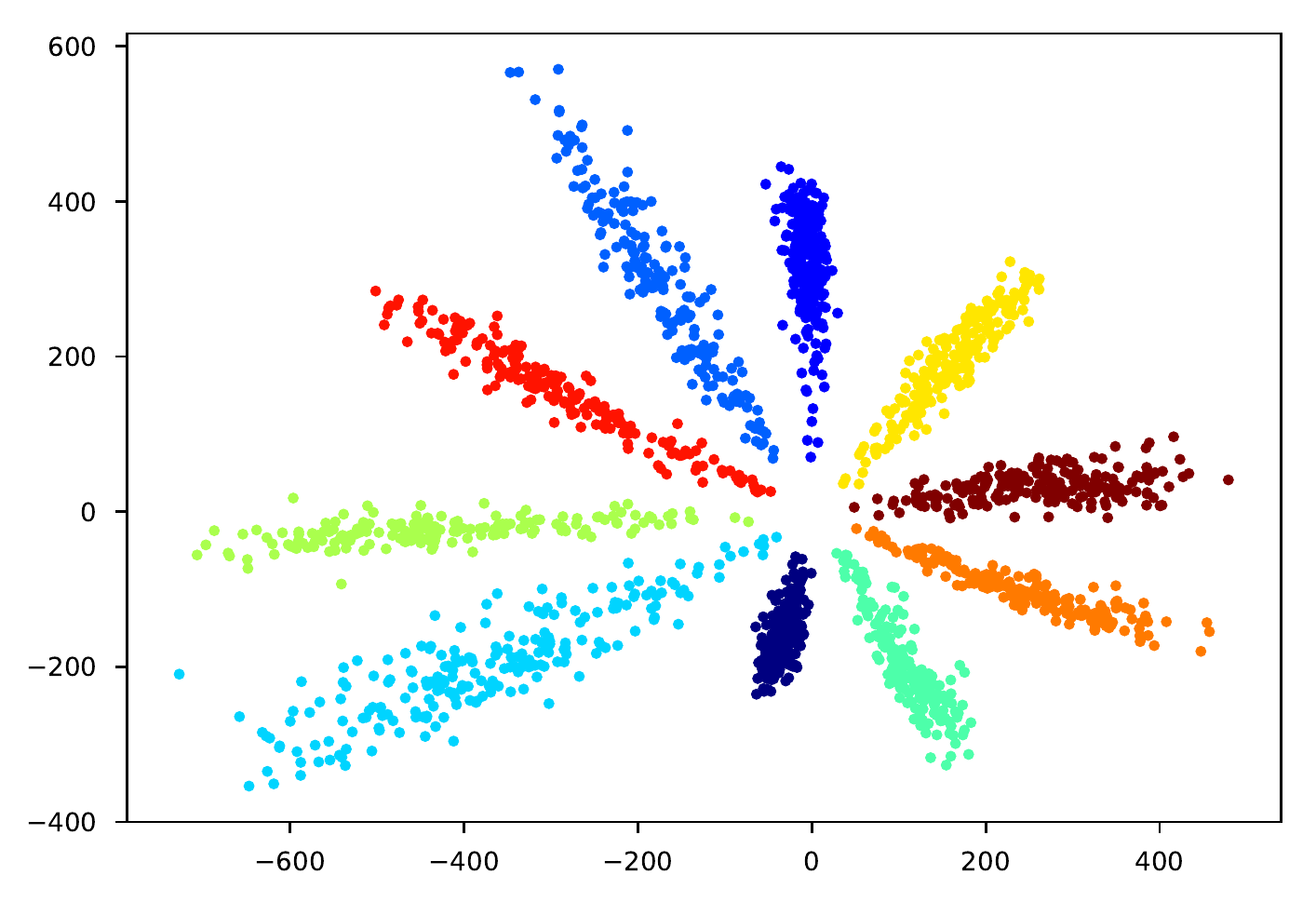}
    \caption{$\mathcal{L}_s$ (2D)}
    \label{2D1}
  \end{subfigure}
   \begin{subfigure}[b]{0.245\textwidth}
    \includegraphics[width=\textwidth]{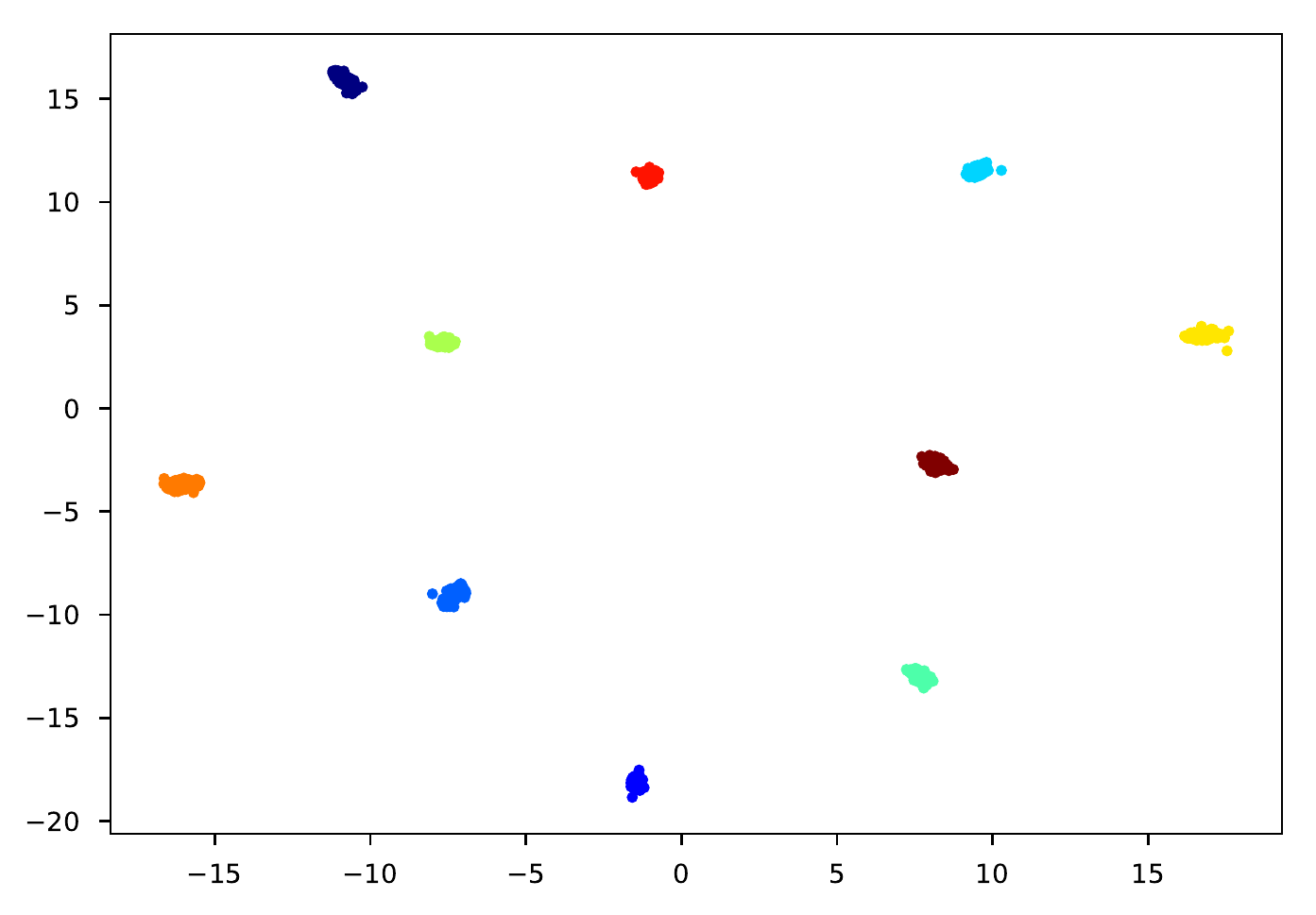}
    \caption{$\mathcal{L}_s$ + $\mathcal{L}_d^{intra}$ (2D)}
    \label{2D2}
  \end{subfigure}
    \begin{subfigure}[b]{0.245\textwidth}
    \includegraphics[width=\textwidth]{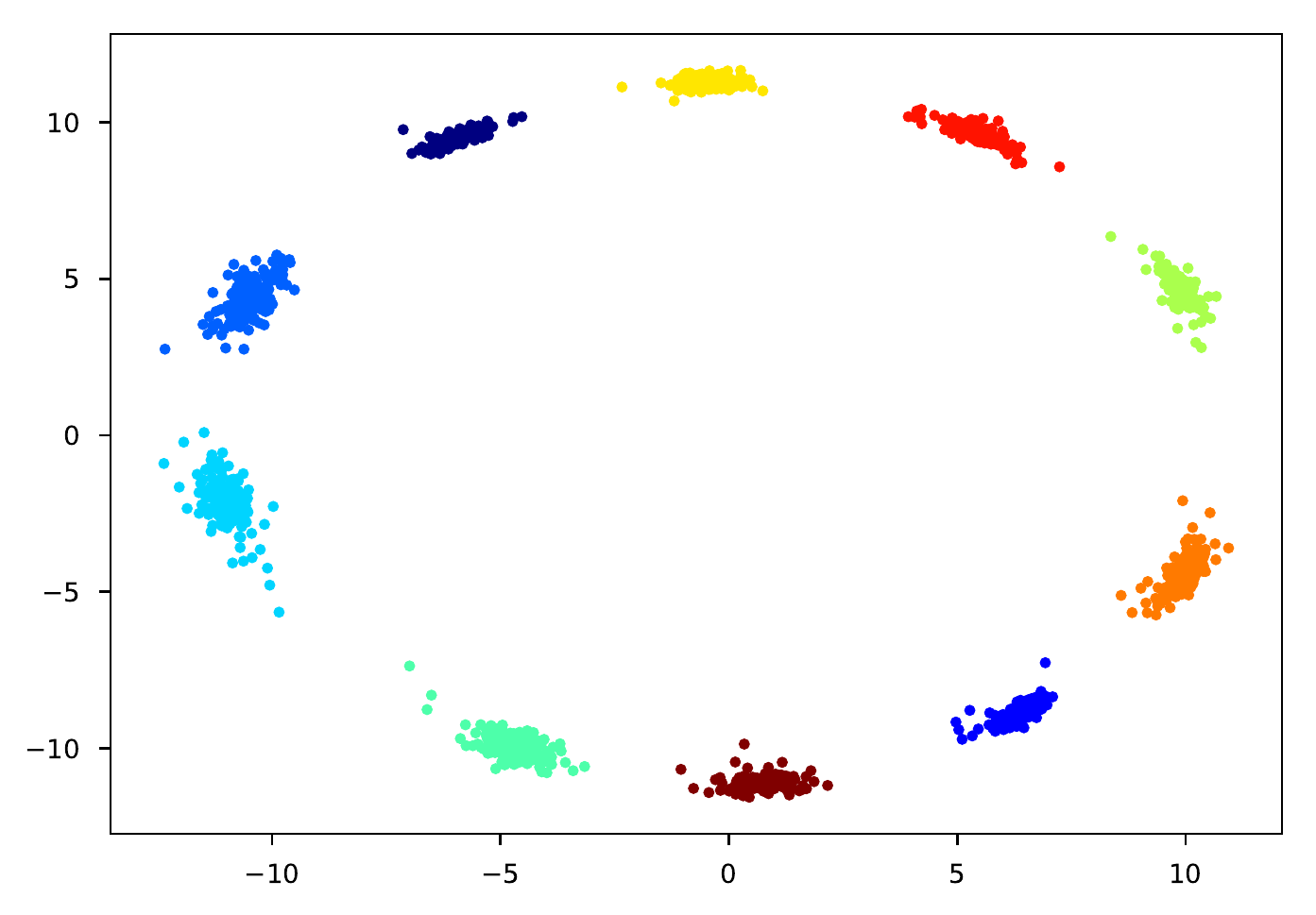}
    \caption{$\mathcal{L}_s$ + $\mathcal{L}_d^{inter}$ (2D)}
    \label{2D3}
  \end{subfigure}
    \begin{subfigure}[b]{0.245\textwidth}
    \includegraphics[width=\textwidth]{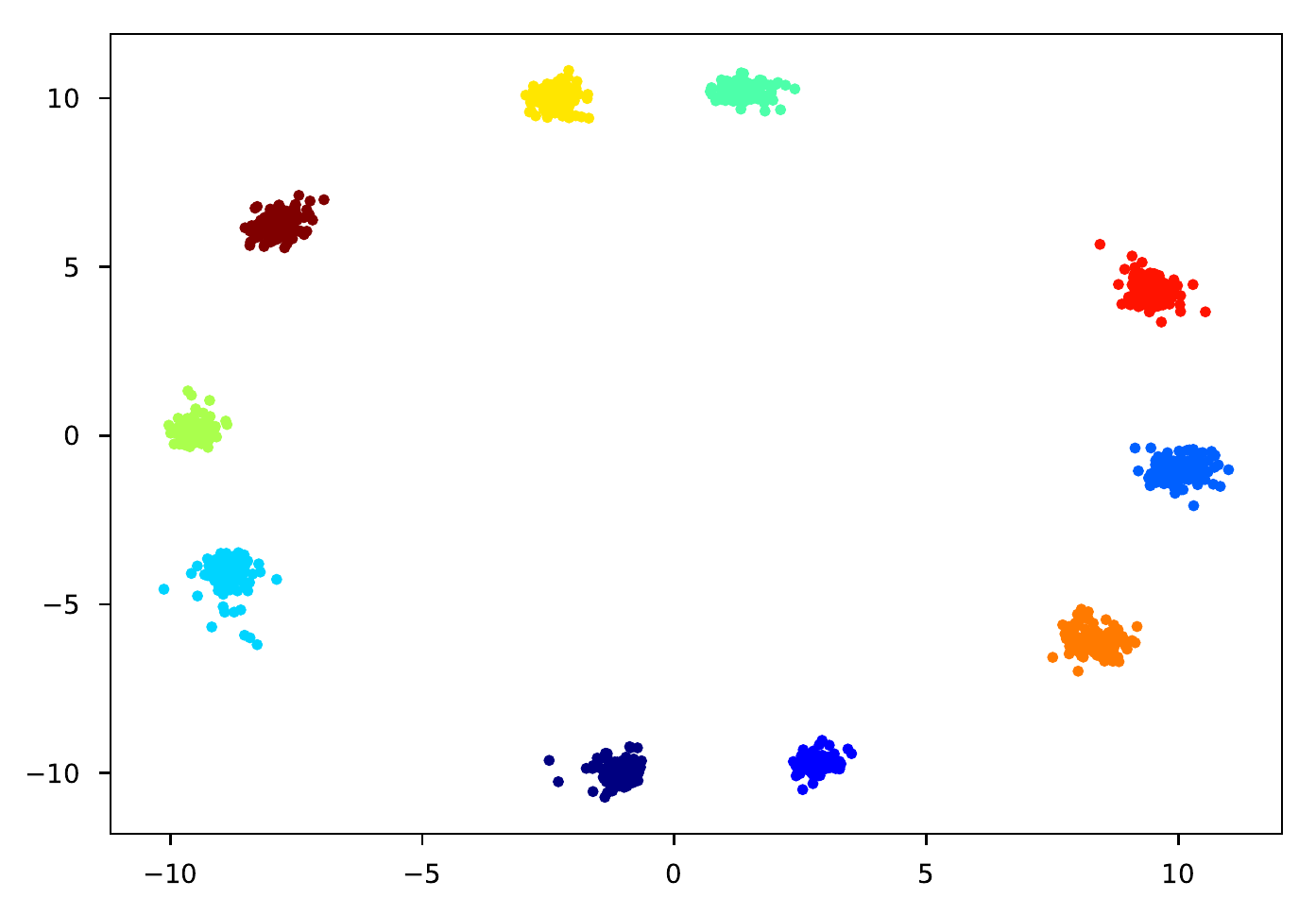}
    \caption{$\mathcal{L}_s$ + $\mathcal{L}_d^{intra}$ + $\mathcal{L}_d^{inter}$ (2D)}
    \label{2D4}
  \end{subfigure}
\caption{ 2D visualization of the deep features. The model is trained (a) with the source loss, (b) with the source loss and the intra-class compactness loss, (c) with the source loss and the inter-class separability loss and (d) with the source loss and the full discrimination loss. It is worth noting that we set the feature dimension in the last FC layer as 2, and then illustrate them by class information.}
\label{fig5}
\end{figure*}

\begin{figure*}[!ht]
\centering
  \begin{subfigure}[b]{0.24\textwidth}
    \includegraphics[width=\textwidth]{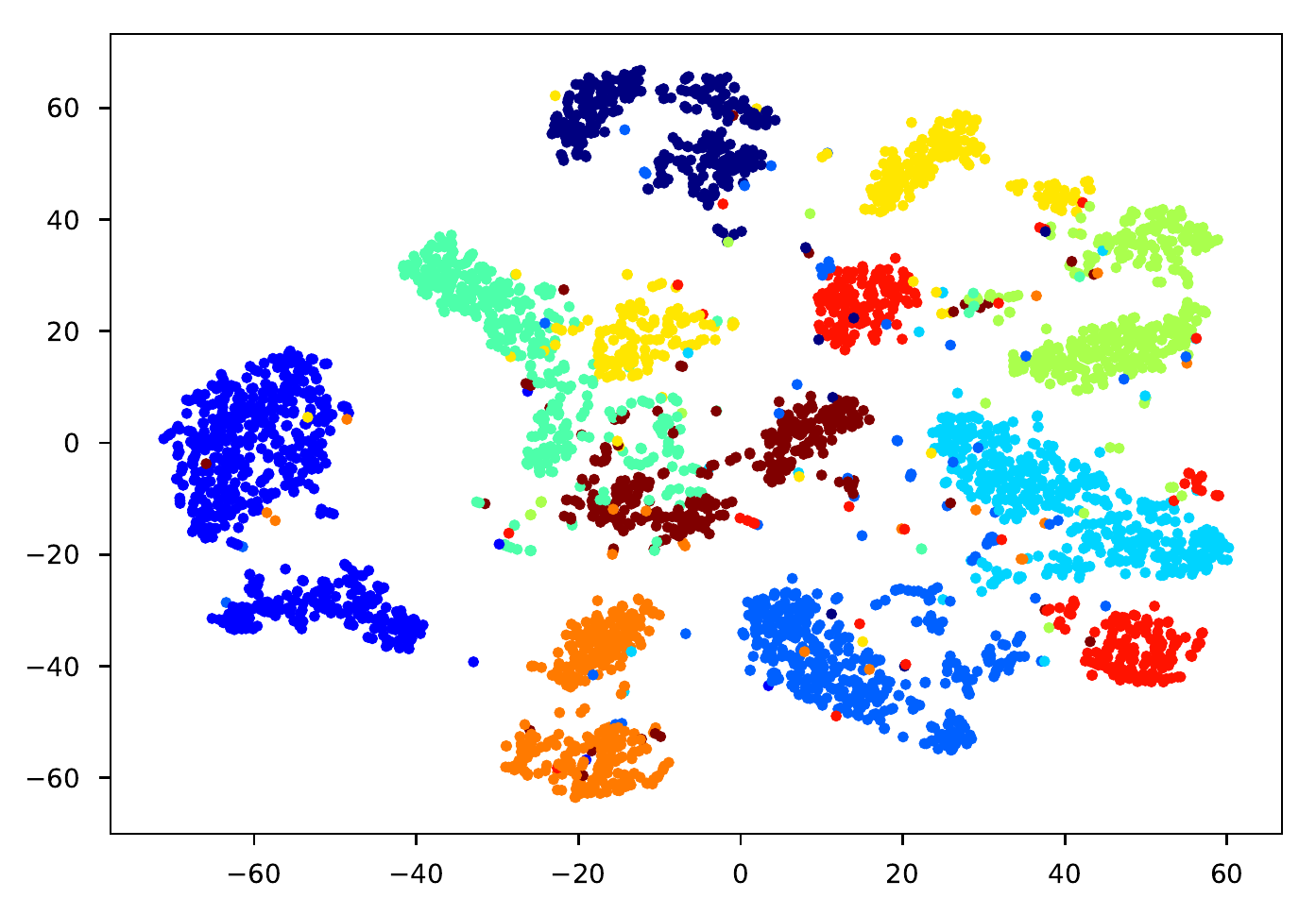}
    \caption{$\mathcal{L}_s$}
    \label{fig:1}
  \end{subfigure}
    \begin{subfigure}[b]{0.24\textwidth}
    \includegraphics[width=\textwidth]{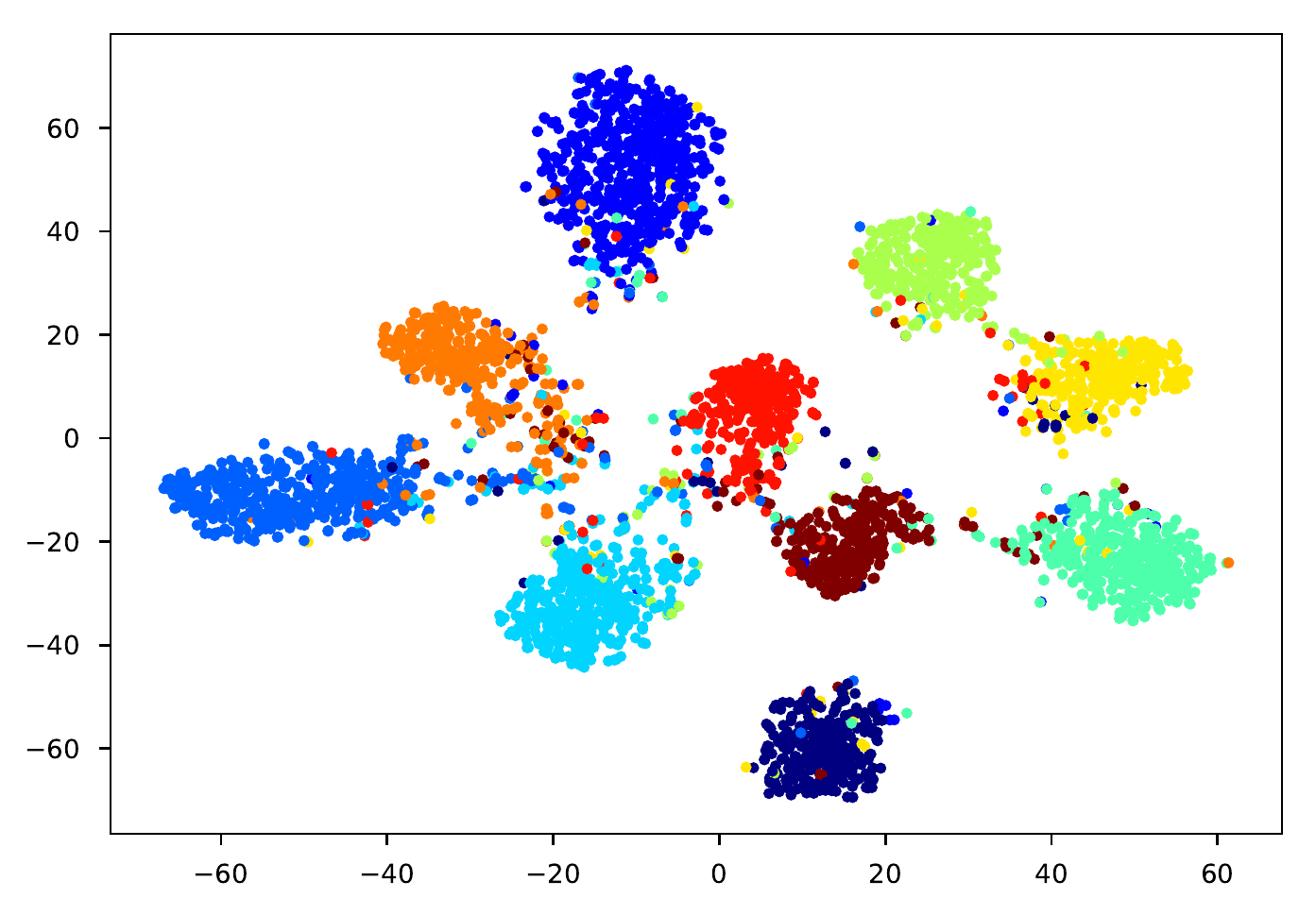}
    \caption{$\mathcal{L}_s$+$\mathcal{L}_{coral}$}
    \label{fig:2}
  \end{subfigure}
  \begin{subfigure}[b]{0.24\textwidth}
    \includegraphics[width=\textwidth]{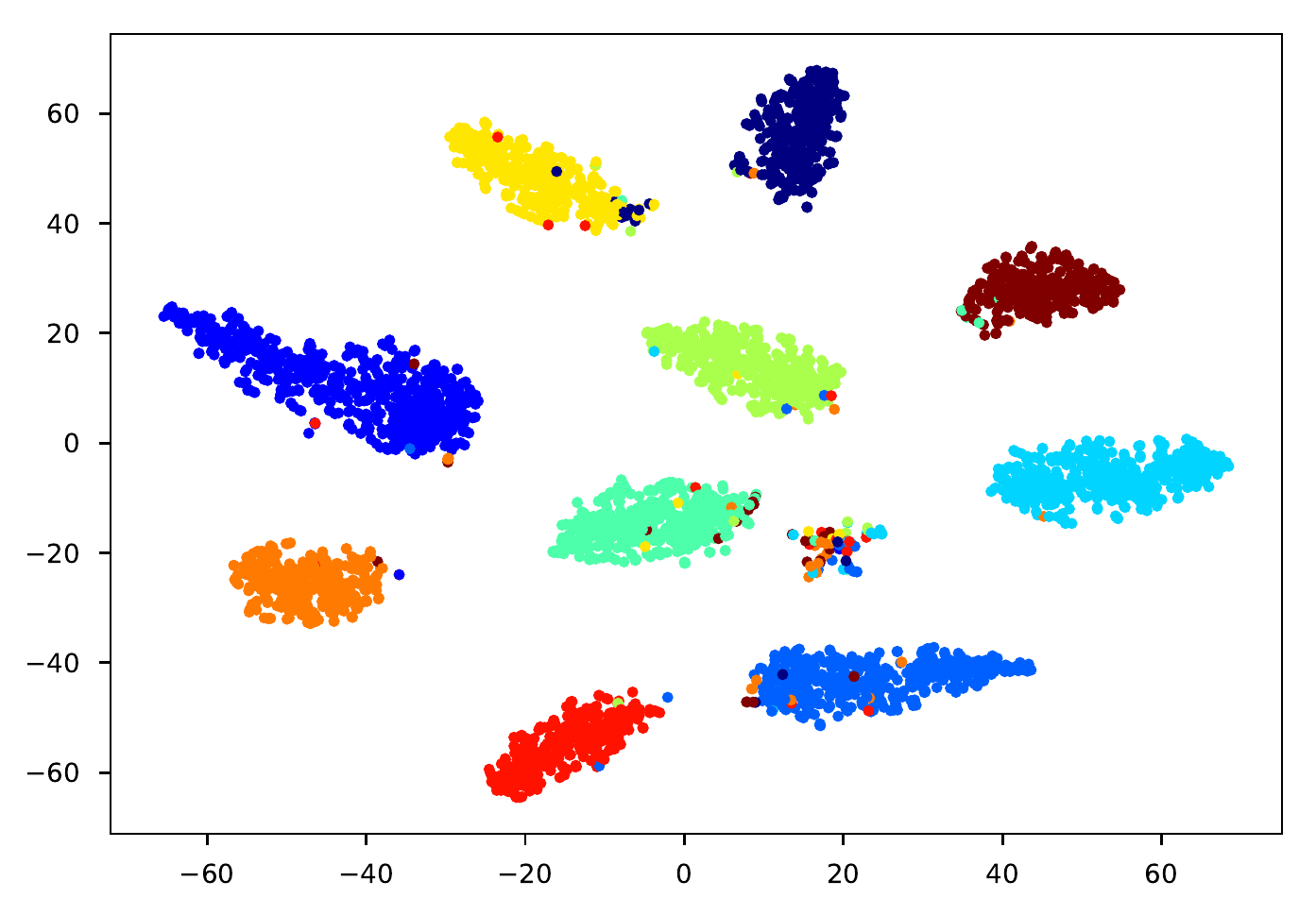}
    \caption{$\mathcal{L}_s$ + $\mathcal{L}_{ssc}$}
    \label{fig:3}
  \end{subfigure}
    \begin{subfigure}[b]{0.24\textwidth}
    \includegraphics[width=\textwidth]{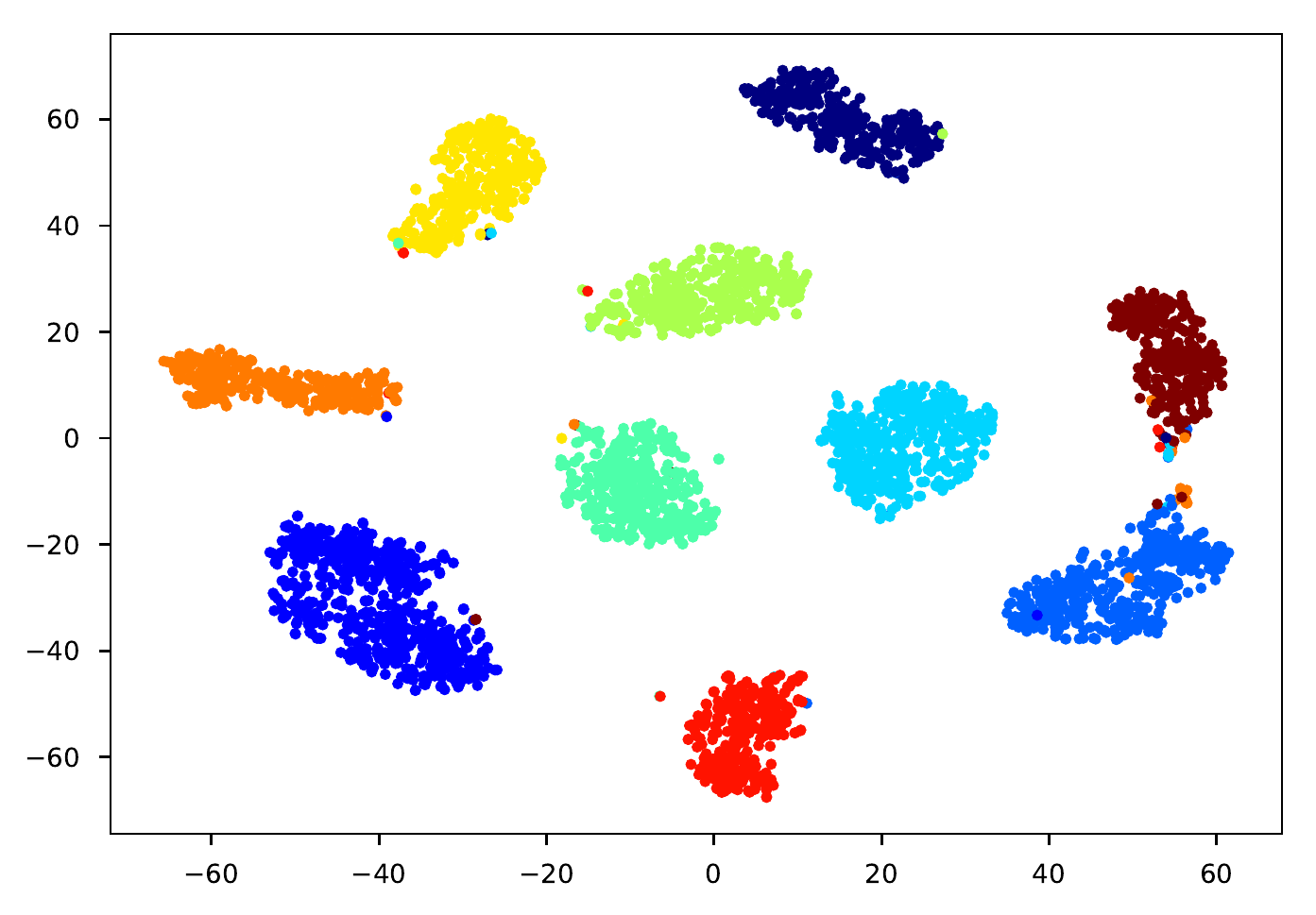}
    \caption{$\mathcal{L}_s$ + $\mathcal{L}_{ssc}$+$\mathcal{L}_d$}
    \label{fig:4}
  \end{subfigure}
  \begin{subfigure}[b]{0.24\textwidth}
    \includegraphics[width=\textwidth]{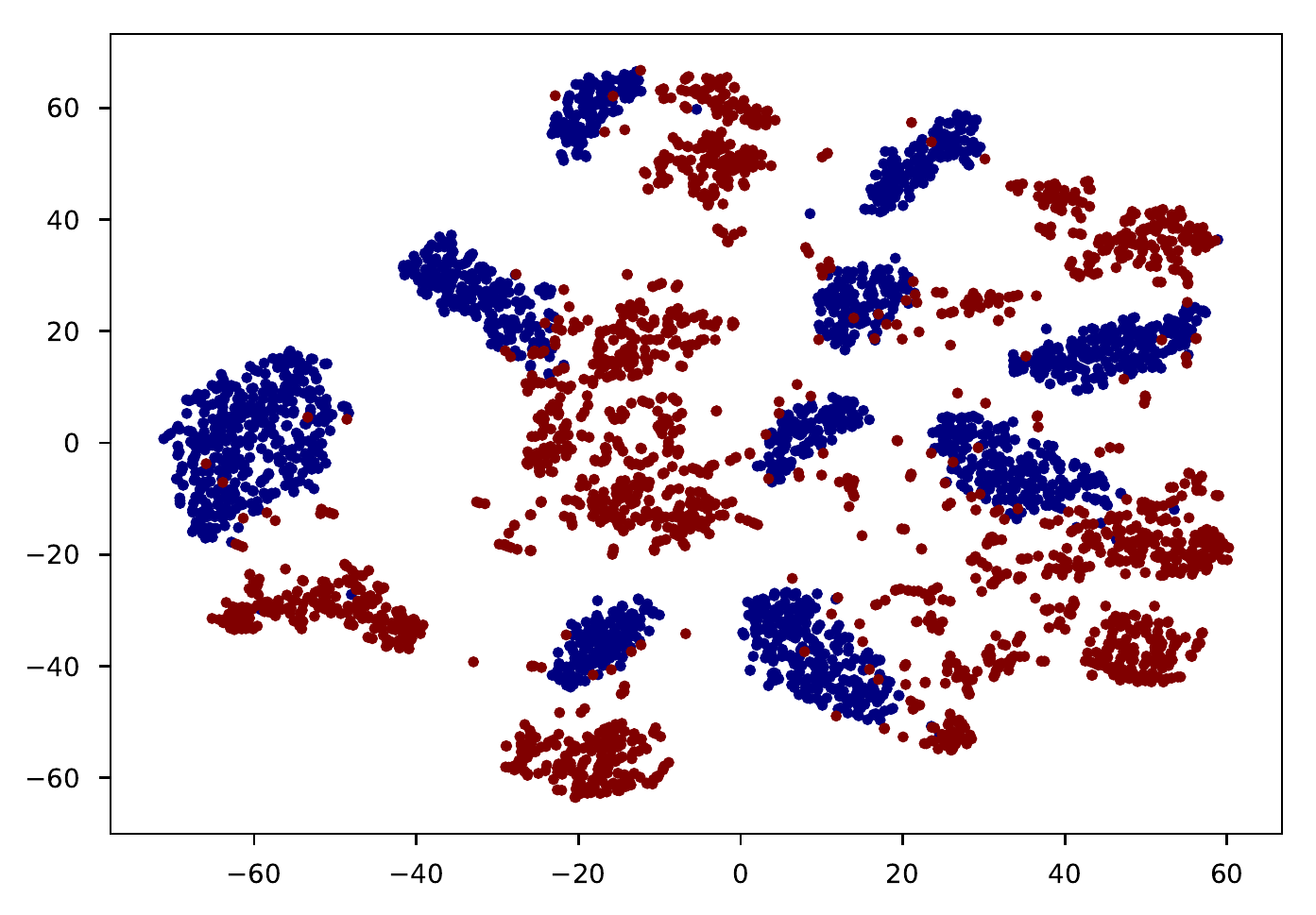}
    \caption{$\mathcal{L}_s$}
    \label{fig:5}
  \end{subfigure}
    \begin{subfigure}[b]{0.24\textwidth}
    \includegraphics[width=\textwidth]{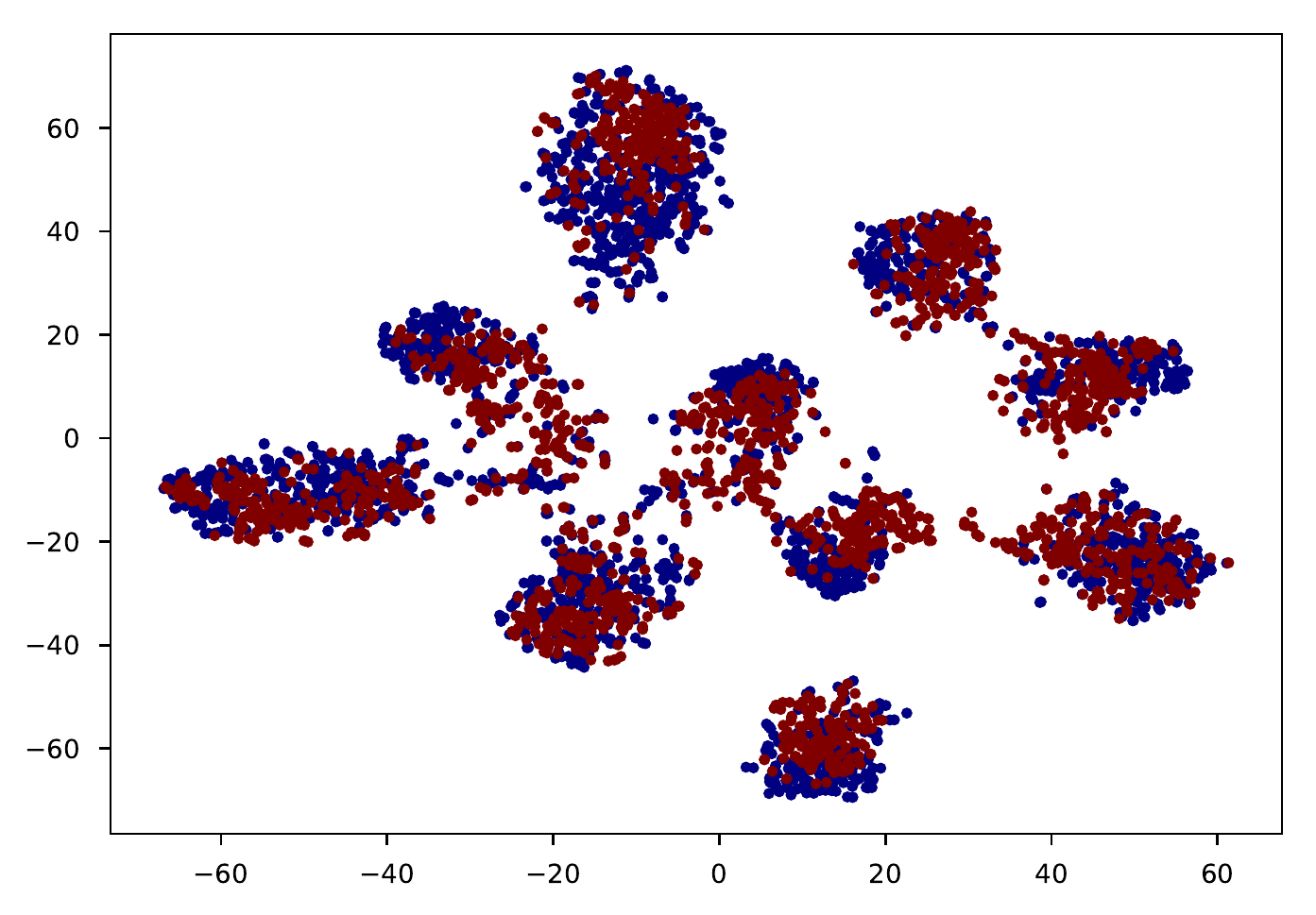}
    \caption{$\mathcal{L}_s$+$\mathcal{L}_{coral}$}
    \label{fig:6}
  \end{subfigure}
  \begin{subfigure}[b]{0.24\textwidth}
    \includegraphics[width=\textwidth]{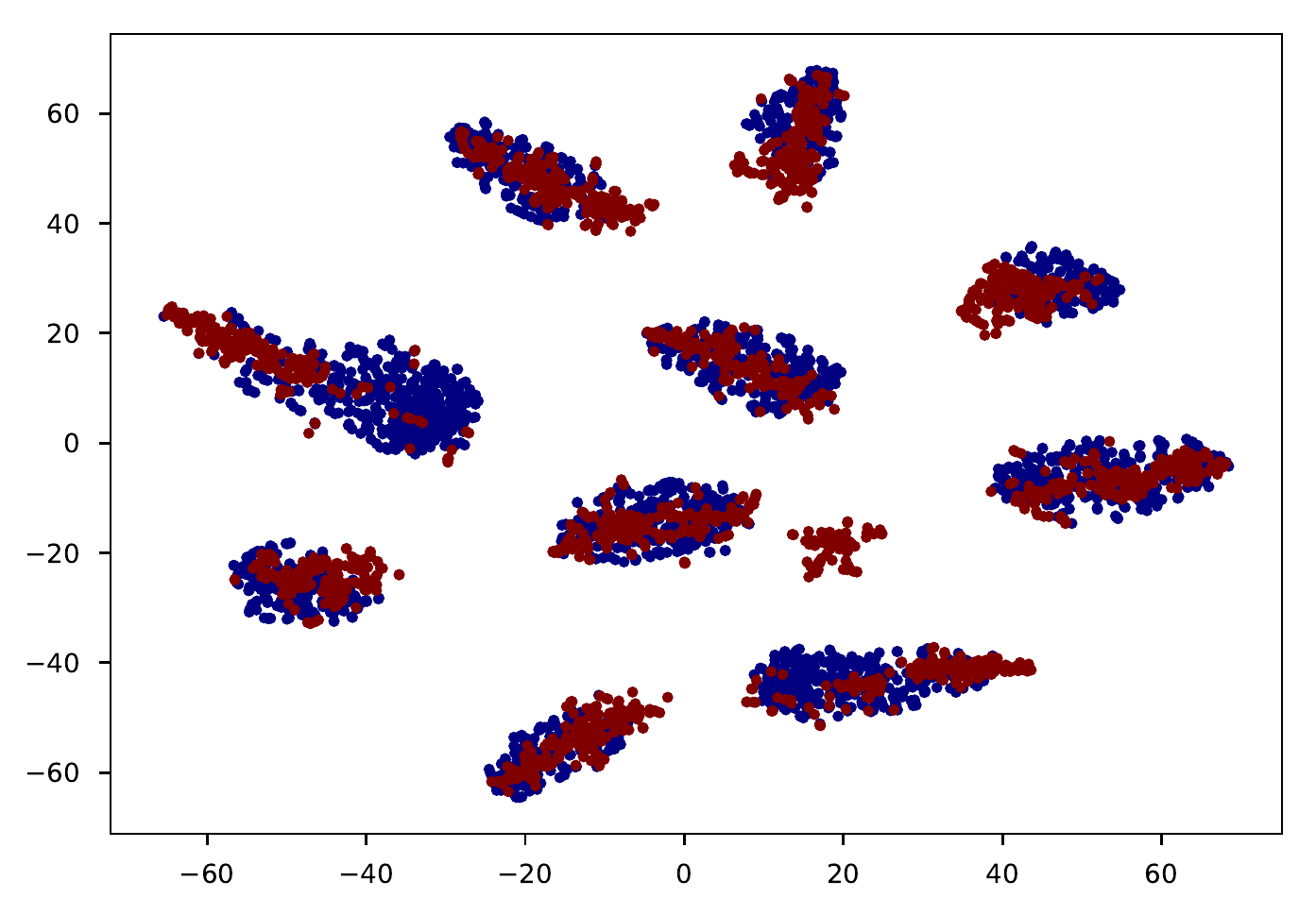}
    \caption{$\mathcal{L}_s$ + $\mathcal{L}_{ssc}$}
    \label{fig:7}
  \end{subfigure}
    \begin{subfigure}[b]{0.24\textwidth}
    \includegraphics[width=\textwidth]{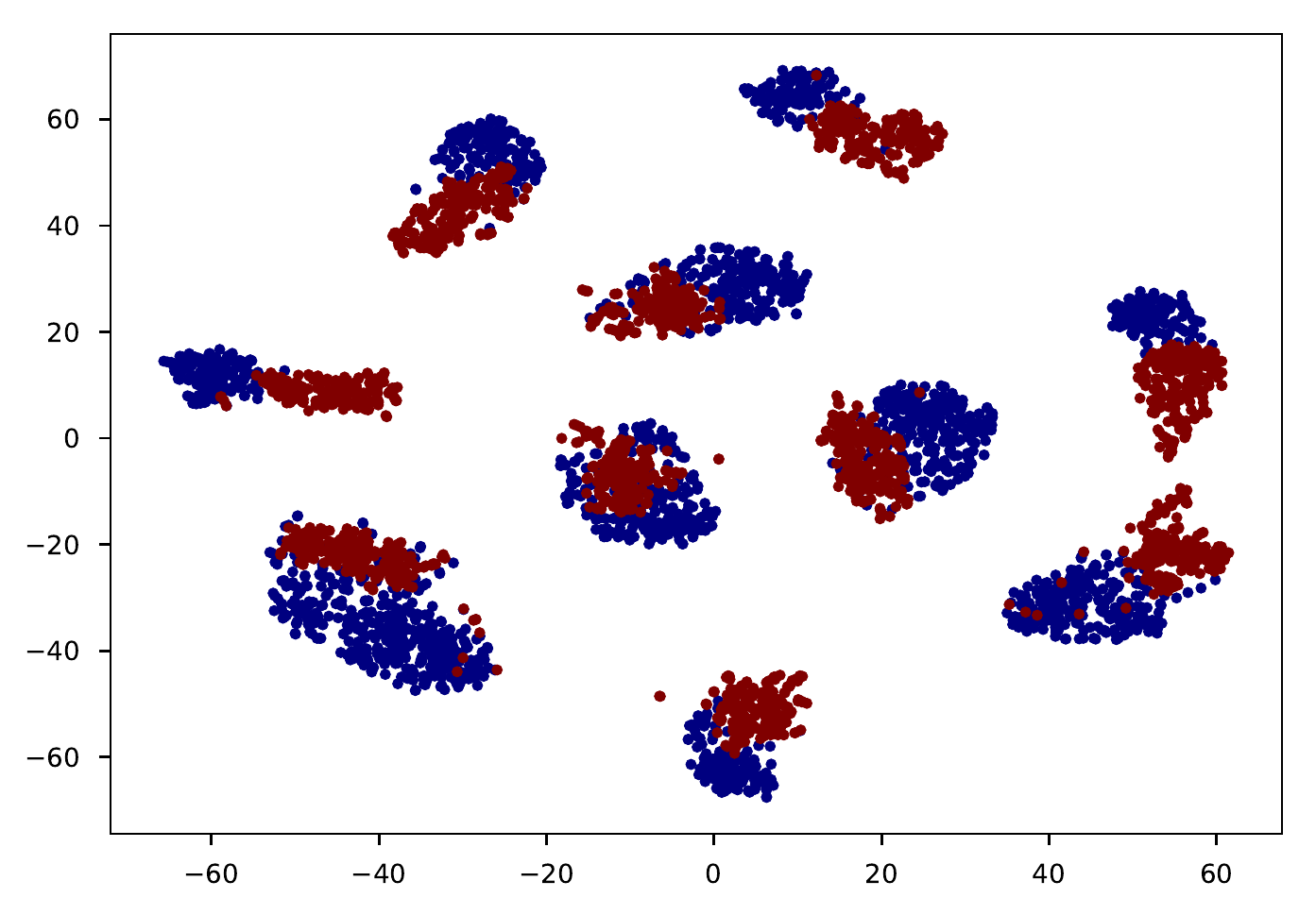}
    \caption{$\mathcal{L}_s$ + $\mathcal{L}_{ssc}$++ $\mathcal{L}_d$}
    \label{fig:8}
  \end{subfigure}
\caption{The t-SNE visualization of the SVHN$\rightarrow$MNIST task. The first row illustrates the t-SNE embedding of deep features which are marked by category information, each color represents a category. The second row illustrates the t-SNE embedding of deep features which are marked by domain information, red and blue points represent the samples drawn from the source and target domains.}
\label{fig6}
\end{figure*}

\textbf{Office-31} The results on Office-31 dataset  for unsupervised domain adaptation based on ResNet-50 are shown in Table \ref{tab2}. It can be seen that our proposal outperforms all the competing methods over all the six transfer tasks. Moreover, our approach improves the classification accuracy substantially on four transfer tasks: A$\rightarrow$W, D$\rightarrow$W, A$\rightarrow$D and D$\rightarrow$A. It is noteworthy that we haven't provided the performance of ADDA on the office-31 dataset. This is because we got very bad results on some transfer tasks, which may be caused by insufficient training data or inappropriate parameter setting. To avoid misleading readers, we didn't write the bad results into the paper.

From Table \ref{tab1} and Table \ref{tab2}, we have several observations: (1) Both the discrepancy-based methods \cite{long2015learning,sun2016deep,zellinger2017central} and the adversarial-based methods \cite{ganin2016domain,tzeng2017adversarial} outperform the source only model by a large margin, which verifies the effectiveness of the deep domain adaptation methods. (2) Our proposed SSC metric outperforms other typical discrepancy metrics, such as MMD, CORAL and CMD on most of the transfer tasks. This can be reflected by comparing the performance of SDDA (w/o FD) and other representative methods. (3) The performance of SDDA (Full) distinctly outperforms the SDDA (w/o FD), which indicates that the proposed feature discrimination constraint can significantly improve the adaptation performance. (4) Compared with the intra-class compactness loss, the introducing inter-class separability loss is more effective to improve the transfer accuracy, which can be demonstrated by comparing the transfer accuracy of SDDA ($\lambda_{inter}=0$) and SDDA ($\lambda_{intra}=0$). (5) The pixel level domain adaptation method CyCADA \cite{hoffman2018cycada} performs well on the digit recognition dataset, especially on the domain shift of MNIST$\rightarrow$MNIST-M, which indicates that CyCADA is more effective for eliminating pixel level and low level domain discrepancy.

\begin{figure*}
    \centering
  \begin{subfigure}[b]{0.23\textwidth}
    \includegraphics[width=\textwidth]{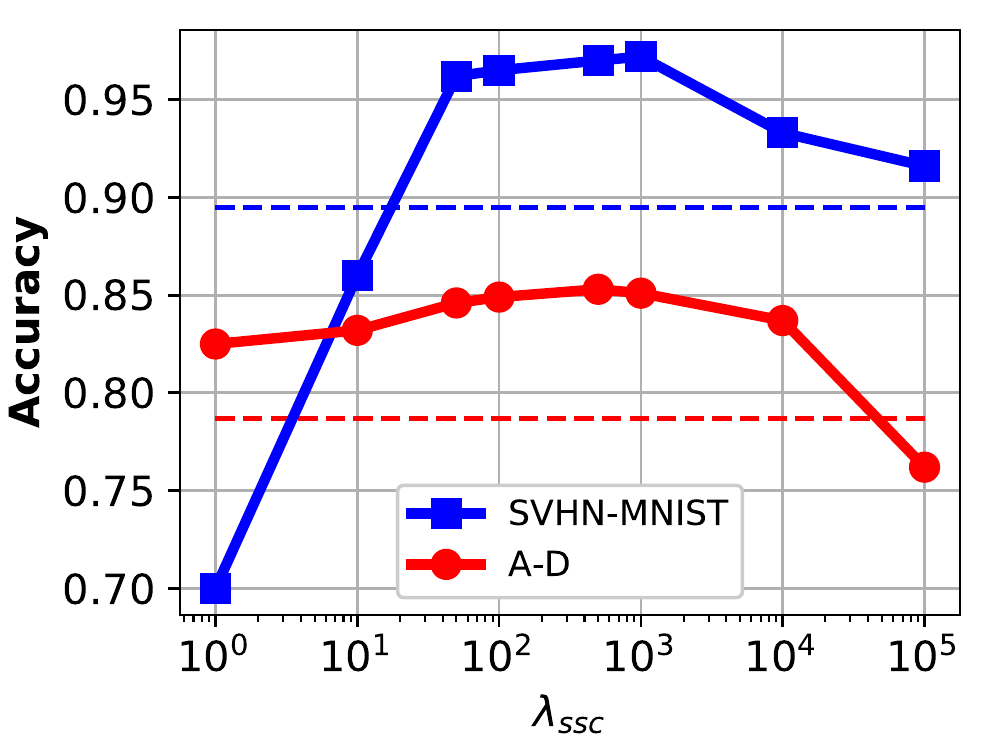}
    \caption{Accuracy w.r.t. $\lambda_{ssc}$}
    \label{sen1}
  \end{subfigure}
   \begin{subfigure}[b]{0.23\textwidth}
    \includegraphics[width=\textwidth]{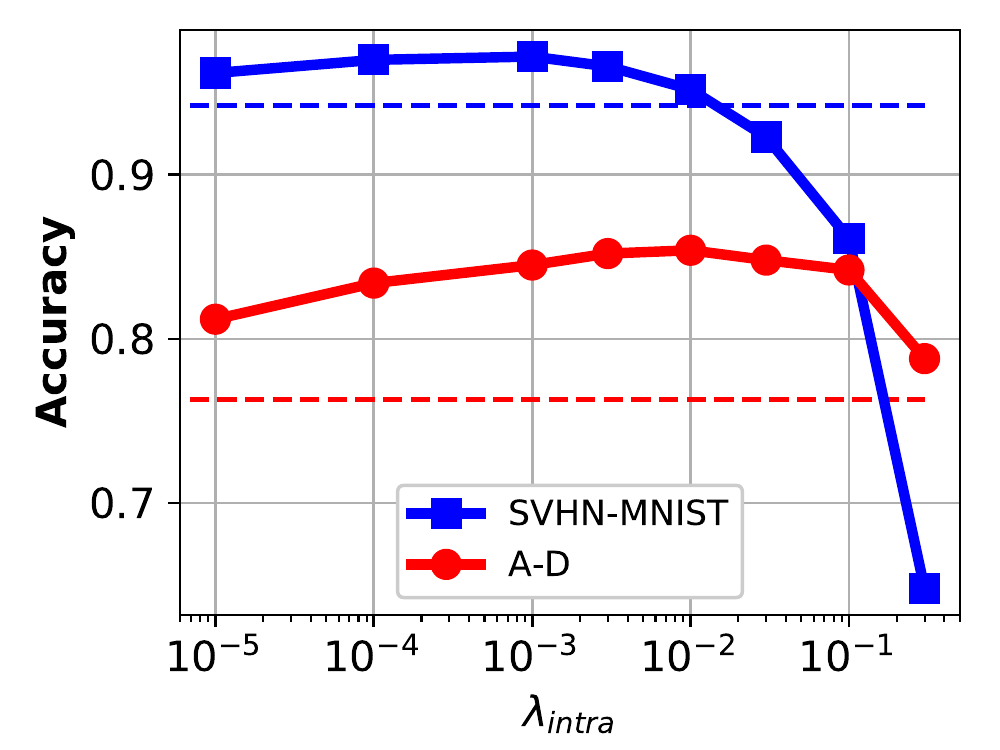}
    \caption{Accuracy w.r.t. $\lambda_{intra}$}
    \label{sen2}
  \end{subfigure}
    \begin{subfigure}[b]{0.23\textwidth}
    \includegraphics[width=\textwidth]{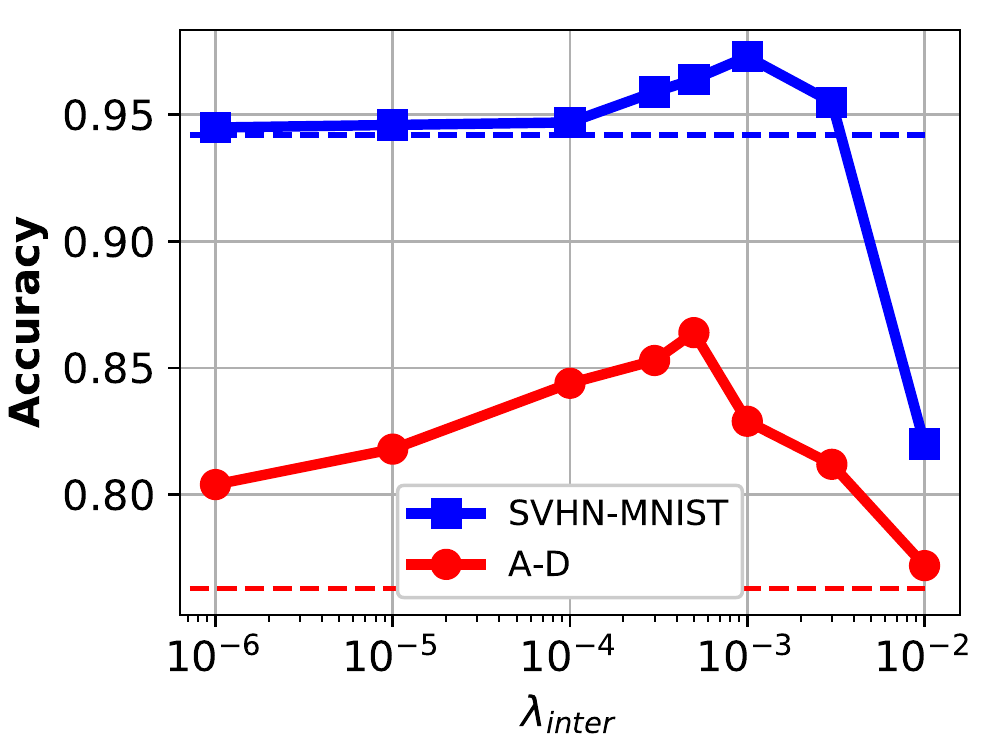}
    \caption{Accuracy w.r.t. $\lambda_{inter}$}
    \label{sen3}
  \end{subfigure}
    \begin{subfigure}[b]{0.23\textwidth}
    \includegraphics[width=\textwidth]{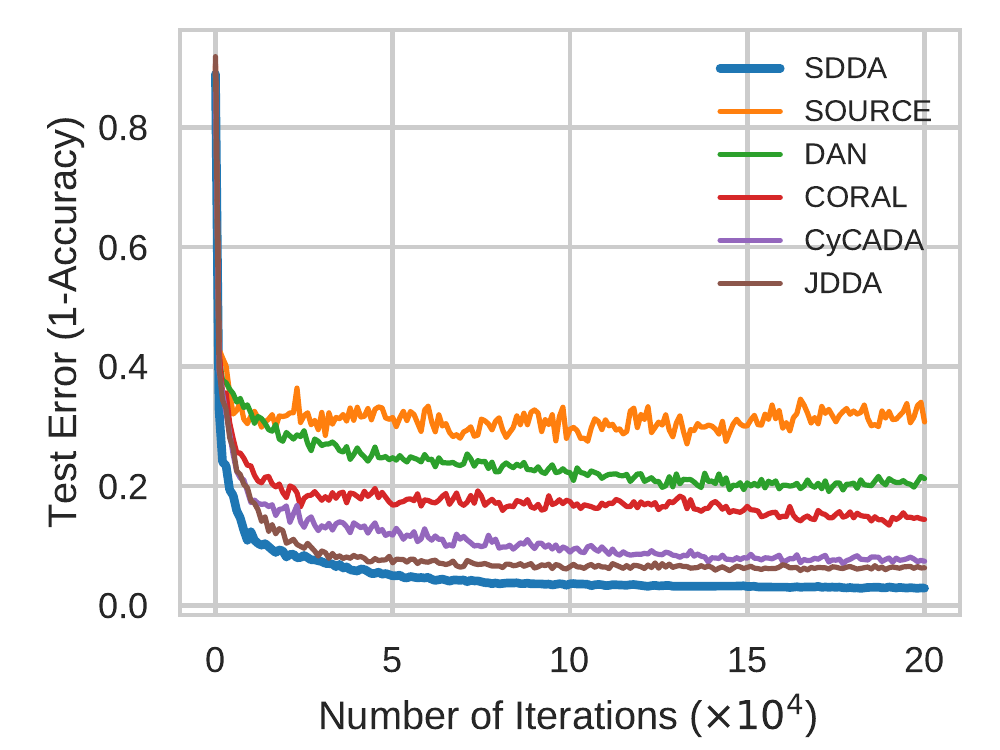}
    \caption{Convergence}
    \label{con1}
  \end{subfigure}
\caption{Parameter sensitivity and convergency analysis. (a-c) The sensitivity of accuracy to $\lambda_{ssc}$, $\lambda_{intra}$, and $\lambda_{inter}$ respectively. (d) The convergency property of SDDA on SVHN$\rightarrow$MNIST. The dashed lines in (a) represent the performance of CORAL, while the dashed lines in (b-c) represent the performance of SDDA(w/o FD).}
\label{fig7}
\end{figure*}

\subsection{Analysis}
\textbf{Feature Visualization} To verify the effectiveness of the introducing feature discrimination constraint, we set the number of hidden nodes in the last FC layer as 2, and visualize the 2D features of 2000 randomly selected source samples. As illustrated in Fig \ref{fig5}, the features trained only with source loss ($\mathcal{L}_s$) follow the strip distribution, while the features trained with source loss and intra-class discrepancy loss ($\mathcal{L}_s$+$\mathcal{L}_d^{intra}$) are tightly clustered. Besides, the features trained with source loss and inter-class discrepancy loss ($\mathcal{L}_s$+$\mathcal{L}_d^{inter}$) are distributed around a circumference, while the features trained with the full discrimination loss ($\mathcal{L}_s$+$\mathcal{L}_d^{intra}$+$\mathcal{L}_d^{inter}$) are well clustered and distributed around a circumference as well. This demonstrates the effectiveness of the introducing feature discrimination loss. We can draw a conclusion that with the feature discrimination constraint, the deep features will be pulled away from the hypersphere origin and be well clustered, which can benefit the domain adaptation tasks.

The visualization of the 2D features only shows the effectiveness of the feature discrimination. Here, we also visualize the t-SNE \cite{maaten2008visualizing} embedding of the learned deep features to demonstrate the effectiveness of our approach visually. As can be seen in Fig. \ref{fig6},  we choose a representative domain shift SVHN$\rightarrow$MNIST with different loss for visualization. From the visualizations, we can make several intuitive observations: (1) The feature distributions of the source only model (non-adapted model) in Fig. \ref{fig:1} and Fig. \ref{fig:5} suggest that the domain shift between SVNH and MNIST datasets is significant, which demonstrates the necessity of domain adaptation. Besides, Fig. \ref{fig:1} and Fig. \ref{fig:5} also verify our declaration that the deep features are fairly domain invariant, i.e., the samples drawn from the same class across domains are sufficiently close, and the samples drawn from different classes have large enough margins, which suggests the insight of learning discriminative features for domain adaptation. (2) Compared with Fig. \ref{fig:2} and Fig. \ref{fig:6} ($\mathcal{L}_s$+$\mathcal{L}_{coral}$), there are less scatter points distributed in the interval of different classes in Fig. \ref{fig:3} and Fig. \ref{fig:7} ($\mathcal{L}_s$ + $\mathcal{L}_{ssc}$), which demonstrates that our proposed SSC metric is more effective to eliminate the domain discrepancy than CORAL. (3) There are less incorrectly clustered samples in Fig. \ref{fig:4} and Fig. \ref{fig:8} ($\mathcal{L}_s$ + $\mathcal{L}_{ssc}$+$\mathcal{L}_d$) than \ref{fig:3} and Fig. \ref{fig:7} ($\mathcal{L}_s$ + $\mathcal{L}_{ssc}$) which demonstrates the merits of learning discriminative features for domain adaptation. (4) As can be seen in Fig. \ref{fig:4} and Fig. \ref{fig:8}, although the source domain and target domain are not perfectly aligned, there is a distinct gap between different classes which guarantees high transfer accuracy of domain adaptation.

\textbf{Parameter Sensitivity}
We conduct empirical parameter sensitivity on two representative transfer tasks SVHN$\rightarrow$MNIST and A$\rightarrow$D. Three trade-off parameters involved in our approach $\lambda_{ssc}$, $\lambda_{intra}$ and $\lambda_{inter}$ are evaluated. Fig. \ref{sen1} demonstrates the transfer accuracy by varying $\lambda_{ssc}\in\{1,10,50,1e2,5e2,1e3,1e4,1e5\}$,  Fig. \ref{sen2} demonstrates the transfer accuracy by varying $\lambda_{intra}\in\{1e-5, 1e-4,1e-3,3e-3,0.01,0.03,0.1,0.3\}$, while Fig. \ref{sen3} demonstrates the transfer accuracy by varying $\lambda_{inter}\in\{1e-6, 1e-5,1e-4,3e-4,5e-4,1e-3,3e-3,1e-2\}$. We can observe that the accuracy of SDDA first increases and then decreases and shows a bell-shaped curve in all the three illustrations, which confirms the effectiveness of the proposed SSC metric and feature discrimination constraint. It is worth noting that since the different distributions of samples in different datasets, the optimal trade-off parameters are distinct on different transfer tasks. The reasonable choice can be $\lambda_{ssc}\in[100,10000]$, $\lambda_{intra}\in[0.0001,0.1]$ and $\lambda_{inter}\in[0.0001,0.005]$. In addition, we can make the conclusion that the model performance is more sensitivity to $\lambda_{inter}$ compared with $\lambda_{intra}$. It also reflects that enhancing the inter-class separability is more effective to improve the feature discrimination and boost the adaptation performance.

\textbf{Convergence Performance}
We also evaluate the convergence performance of our proposal though the test error during the training phase. Fig. \ref{con1} shows the test errors of different methods on the transfer tasks SVHN$\rightarrow$MNIST. It suggests that our proposed SDDA performs best and converges fastest compared with other state-of-the-art methods, which reveals the effectiveness of our proposal.

\section{Conclusions}
To minimize the domain discrepancy for cross-domain knowledge transfer, we propose to match the structure of the deep features in the source and target domains by a self-similarity consistency constraint, instead of aligning the global distribution statistics across domains. The experimental analysis demonstrates the superiority of the proposed SSC metric compared with the widely used MMD and CORAL. Besides, we also propose to improve the adaptation performance by learning more discriminative features when perform domain matching. An elegant feature norm constraint is exploited to enlarge the margins of inter-class samples. Experimental results and feature visualization suggest that learning more discriminative features can ease the requirements of strict domain alignment and improve the transfer accuracy effectively. The source code of this work will be released soon.

%
\bibliographystyle{ACM-Reference-Format}
\bibliography{sample-base.bib}
\end{document}